\documentclass[journal]{IEEEtran}
\usepackage{cite}

\ifCLASSINFOpdf
   \usepackage[pdftex]{graphicx}
  \graphicspath{{fig/}}
   \DeclareGraphicsExtensions{.pdf,.jpeg,.png,.jpg}
\else
   \usepackage[dvips]{graphicx}
   \graphicspath{{../pdf/}}
   \DeclareGraphicsExtensions{.eps}
\fi
\usepackage{amsmath}
\usepackage{amsfonts}
\ifCLASSOPTIONcompsoc
  \usepackage[caption=false,font=normalsize,labelfont=sf,textfont=sf]{subfig}
\else
  \usepackage[caption=false,font=footnotesize]{subfig}
\fi
\usepackage{fixltx2e}

\usepackage{color}
\definecolor{deemph}{gray}{0.6}

\usepackage{subfig}

\def\onedot{\ifx\@let@token.\else.\null\fi\xspace}

\newcommand\etc{\emph{etc.}}

\makeatother
\def\Vec#1{{\boldsymbol{#1}}}
\def\Mat#1{{\boldsymbol{#1}}}
\newcommand\etal{\emph{et al.}}

\usepackage[thmmarks,amsmath]{ntheorem}
\newtheorem{remark}{Remark}

\ifCLASSOPTIONcaptionsoff
  \usepackage[nomarkers]{endfloat}
 \let\MYoriglatexcaption\caption
 \renewcommand{\caption}[2][\relax]{\MYoriglatexcaption[#2]{#2}}
\fi

\usepackage{marvosym}
\usepackage{ifsym}

\usepackage{rotating}
\usepackage{multirow}
\usepackage[pagebackref=false,breaklinks=true,letterpaper=true,colorlinks,bookmarks=false]{hyperref}

\begin{document}
\title{Asymmetric Dual-Decoder U-Net for Joint Rain and Haze Removal}

\author{Yuan~Feng$^*$,~\IEEEmembership{Member, IEEE},
        Yaojun~Hu$^*$,
        Pengfei~Fang$\textsuperscript{\Letter}$,
        Yanhong~Yang,
        Sheng~Liu and
        Shengyong~Chen,~\IEEEmembership{Senior Member, IEEE}
        
\thanks{This work was supported by Zhejiang Provincial Natural Science Foundation of China (LGG21F030011), and National Natural Science Foundation of China (61972355).}

\thanks{Y. Feng and Y. Hu are with the College of Science, Zhejiang University of Technology, Hangzhou 310023, China (e-mail: fy@ieee.org; huuyjun@gmail.com).}

\thanks{P. Fang is with College of Engineering and Computer Science, the Australian National University, Canberra, ACT 2601, Australia (e-mail: Pengfei.Fang@anu.edu.au).}

\thanks{S. Liu is with the College of Computer Science, Zhejiang University of Technology, Hangzhou 310023, China (e-mail: edliu@zjut.edu.cn).}

\thanks{Y. Yang and S. Chen are with the College of Computer Science and Technology, Tianjin University of Technology, Tianjin 300384, China. (e-mail: yyh\_03@163.com; sy@ieee.org).}
\thanks{$*$ Equal contribution.}
\thanks{$\text{\Letter}$ Corresponding author.}
}

%
%

\markboth{Journal of \LaTeX\ Class Files,~Vol.~14, No.~8, August~2015}%
{Shell \MakeLowercase{\textit{et al.}}: Bare Demo of IEEEtran.cls for IEEE Journals}
%



\maketitle


\begin{abstract}
This work studies the joint rain and haze removal problem. In real-life scenarios, rain and haze, two often co-occurring common weather phenomena, can greatly degrade the clarity and quality of the scene images, leading to a performance drop in the visual applications, such as autonomous driving. However, jointly removing the rain and haze in scene images is ill-posed and challenging, where the existence of haze and rain and the change of atmosphere light, can both degrade the scene information. Current methods focus on the contamination removal part, thus ignoring the restoration of the scene information affected by the change of atmospheric light. We propose a novel deep neural network, named \underline{A}symmetric \underline{D}ual-decoder \underline{U-Net} (ADU-Net), to address the aforementioned challenge. The ADU-Net produces both the contamination residual and the scene residual to efficiently remove the rain and haze while preserving the fidelity of the scene information. Extensive experiments show our work outperforms the existing state-of-the-art methods by a considerable margin in both synthetic data and real-world data benchmarks, including RainCityscapes, BID Rain, and SPA-Data. For instance, we improve the state-of-the-art PSNR value by $2.26$/$4.57$ on the RainCityscapes/SPA-Data, respectively.

Codes will be made available freely to the research community.
\end{abstract}



%

\IEEEpeerreviewmaketitle

\section{Introduction}

\IEEEPARstart{W}{hen} photographing in bad weather, the quality of outdoor scene images can be greatly degraded by the contamination, i.e., rain, haze or snow, etc. distributed in the air. Such contamination absorbs or disperses the scene light, thereby reducing the contrast and color fidelity of the scene image. Hence, the existence of contamination significantly affects many real-world vision systems, such as scene recognition, object tracking, semantic segmentation, etc, and all of these vision systems are essential for autonomous driving~\cite{Chen2019DeepIntergration, fan2019lasot, Zhang2019Co-O}. In another word, such outdoor vision systems, which works efficiently in ideal weather condition, will suffer a plummet due to complex real-world weather conditions. Therefore, it is essential to develop algorithms to restore images contaminated by different contaminants as a pre-processor for such outdoor vision systems.

\begin{figure}
\centering
\subfloat[Input]{
\includegraphics[width=0.32\linewidth]{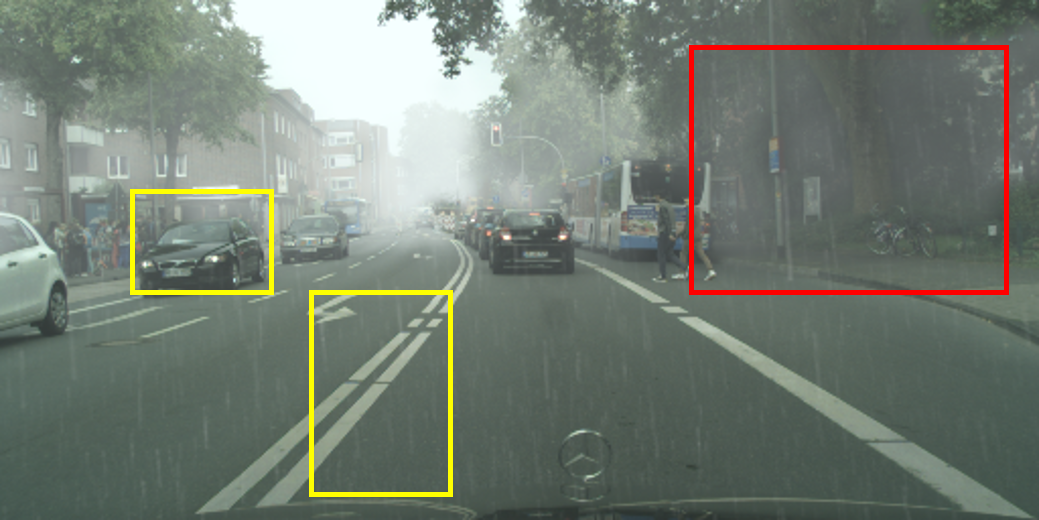}
\label{input}
}
\subfloat[Ground Truth]{
\includegraphics[width=0.32\linewidth]{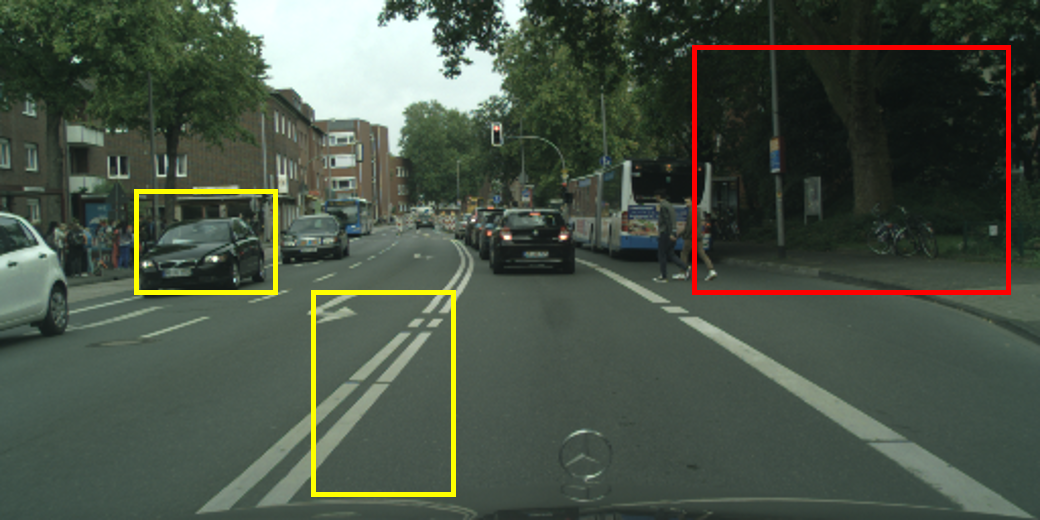}
\label{gt}
}
\subfloat[True Res]{
\includegraphics[width=0.32\linewidth]{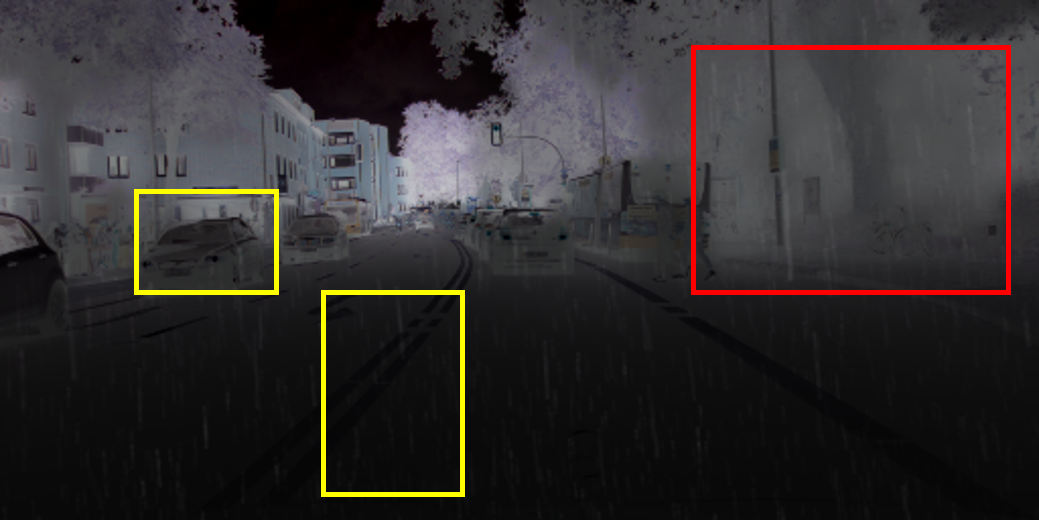}
\label{residualgt}
}
\hfill
\subfloat[Contamination Res]{
\includegraphics[width=0.32\linewidth]{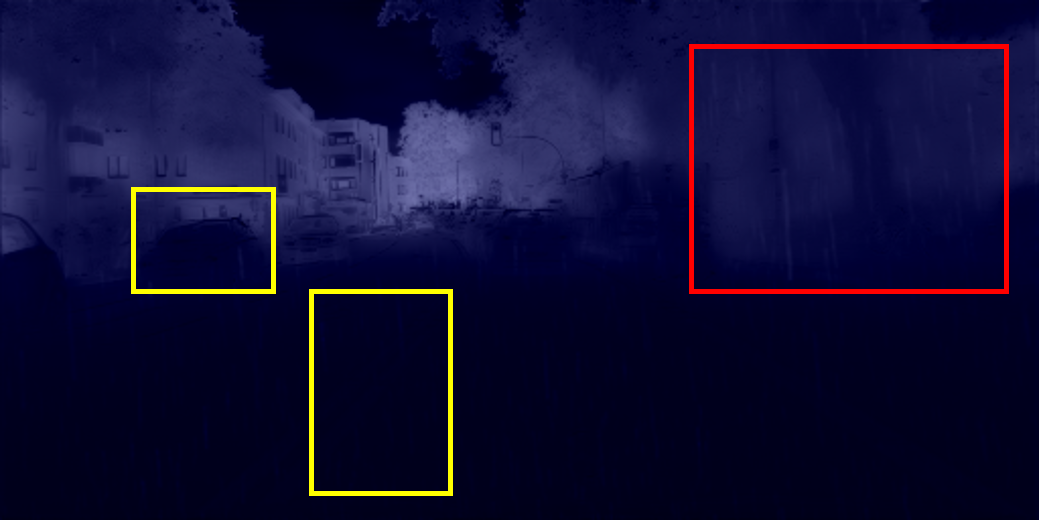}
\label{fore}
}
\subfloat[Scene Res]{
\includegraphics[width=0.32\linewidth]{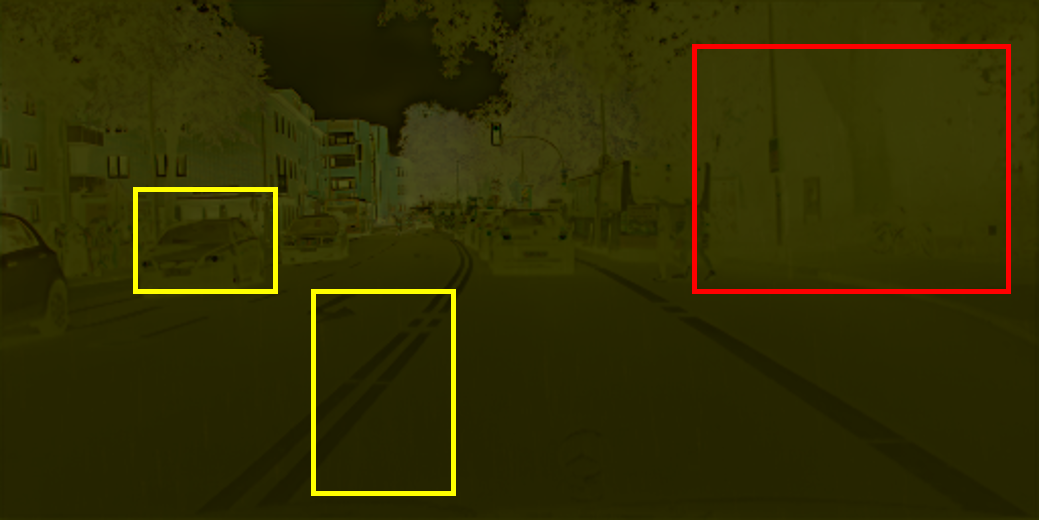}
\label{back}
}
\subfloat[Output Res]{
\includegraphics[width=0.32\linewidth]{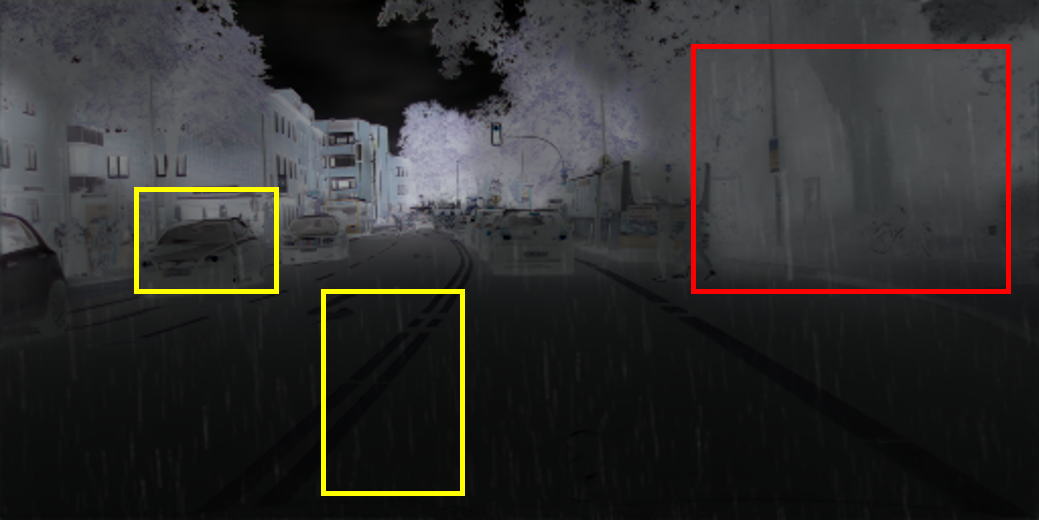}
\label{residualout}
}
\caption{Example of a scene image and its residual maps. (a) is the input image and (b) is the ground truth from RainCityscapes dataset. Image in (c) is the difference between (a) and (b). (d) and (e) are the contamination residual and scene residual. (f) is the result of (a)+(b). ``Res'' indicates ``Residual''. The \textcolor{red}{contamination} and \textcolor{yellow}{scene} details are included in the \textcolor{red}{red} and \textcolor{yellow}{yellow} boxes, respectively (zoom in to find the details).}\label{result}
\end{figure}

In this work, we focus on a real yet less-investigated scenario, the co-occurrence of the rain and haze in the scenes. Both image rain removal and haze removal are challenging low-level computer vision tasks. Many efforts have been made to solve the individual rain removal and haze removal tasks~\cite{rcdnet2020wang,zamir2021multi,AECRNET}. However, only a few works consider removing the rain and haze jointly in scene images~\cite{dgnl2021,FHRR2021,blind2021han}. In the real-world scenario, it is a very common situation that the rain and haze co-occur in the rainfall environment (see Fig.~\ref{input})~\cite{visionandrain}. Along with rain streaks and raindrops, the uneven haze will also obscure the image, interfering with the perception of the environment. Such a scenario brings challenges to the outdoor vision systems that are required to jointly remove the rain and haze in images.

The existing methods for single-image rain and haze removal can be roughly categorized into two categories: priority knowledge-oriented approaches and data-driven approaches. The prior knowledge-based image rain removal~\cite{kang2011automatic,luo2015removing,li2016rain} and haze removal methods~\cite{he2010single,fattal2014dehazing,zhu2015fast} are mostly based on the physical imaging models. However, such solutions suffer from the robustness issue when deployed into real-world scenarios~\cite{zhu2017joint, chinese2021}. Recent advances in deep learning demonstrate dramatic success in haze removal~\cite{li2017aod,ren2020single,chen2019pms} and rain removal~\cite{zhang2019image,ren2019progressive,wang2019spatial}. Learning-based methods in both fields have achieved cutting-edge performance on synthetic datasets. However, methods designed for certain contamination cannot handle the complex real-world scenario with the co-occurrence of the rain and haze in the natural scenes.  Recent studies also pointed out the necessity of joint-removal, such as Han \etal{}~\cite{blind2021han} decompose rain and haze by a Blind Image Decomposition Network, and Kim \etal{}~\cite{FHRR2021} remove rain and haze by a frequency-based model. A new dataset for the purpose of benchmarking joint rain and haze removal, named RainCityScapes, is also proposed to facilitate research on this important task~\cite{dgnl2021}. Thus such a joint-removal task becomes an open problem in the community and calls for further study.

Recent advances in low-level computer vision have made remarkable progress, where a well-trained deep neural network can almost perfectly remove the contamination in the outdoor scene images. However, no existing work considers paying attention to the scene difference in the restoration process. We observe that the true residual, obtained by $(\mathrm{Input} - \mathrm{Ground}~\mathrm{Truth})$ (see Fig.~\ref{residualgt}), contains the scene information. That is, a neural network designed to focus on contamination may suffer from a gap in recovering the scenes. Such a gap motivates us to develop a unified method to remove the contamination and compensate for the scene information in one go. 

In real-world scenarios, the weather condition is complex, that is, different components, such as rain streak and haze, may co-occur in the scenes. The occurrence of some components, i.e., heavy haze, impacts the atmospheric light. As a consequence, the scene information at the photometric level can be degraded. Physically speaking, along with removing contamination in the image, it is also necessary to restore scene information affected by the change of atmospheric light. To address this issue, we proposes a novel dual-branch architecture, called \underline{A}symmetric \underline{D}ual-decoder \underline{U-Net} (ADU-Net). The ADU-Net consists of a single branch encoder and asymmetric dual-branch decoders. In the asymmetric dual-branch architecture, one branch, the contamination residual branch, is designed to remove the contamination (see Fig.~\ref{fore}). Another branch, the scene residual branch, is required to perform the recovery of scene information (see Fig.~\ref{back}). The contamination residual branch, equipped with a novel channel feature fusion (CFF) module and window multi-head self-attention (W-MSA), produces the contamination residual. The scene branch, powered by a novel global channel feature fusion (GCFF) module and shift-window multi-head self-attention (SW-MSA) mechanism, aims to preserve the scene information by the scene residual. The joint efforts of contamination residual and scene residual separate the rain and haze from the input scene image, while preserving the scene of the image (see Fig.~\ref{residualout}). The proposed ADU-Net can effectively remove the different contamination in the images and compensates for the scene information on multiple benchmark datasets, including RainCityscapes~\cite{dgnl2021}, BID rain~\cite{blind2021han} and SPA-Data~\cite{wang2019spatial}. 

Our contribution can be summarized as follows: 
\begin{itemize}
\item We propose a novel yet efficient neural architecture, ADU-Net, to jointly remove rain and haze in scene images.   
\item We present an asymmetric dual-decoder, which removes the contamination while compensating for the scene information of the image. To the best of our knowledge, this is the first work to consider the recovery of scene information in deraining and dehazing tasks. 
\item Extensive experiments, including quantitative studies and qualitative studies, are conducted to evaluate the effectiveness of the ADU-Net. Empirical evaluation shows our method outperforms the current state-of-the-art methods by a considerable margin.
\end{itemize}

\section{Related Work}
\subsection{Single-image Rain Removal}
The very first single-image rain removal methods were based on a priori knowledge. Morphological component analysis (MCA)~\cite{kang2011automatic} employs bilateral filters to extract high-frequency components from rain images, where the high-frequency components are further decomposed into "rain components" and "non-rain components" through dictionary learning and sparse coding. Luo \etal{}~\cite{luo2015removing} proposed a single-image rain removal algorithm based on mutual exclusion dictionary learning. Gaussian mixture model prior knowledge~\cite{li2016rain} was utilized to accommodate multiple orientations and scales of rain streaks. In~\cite{zhu2017joint}, Zhu \etal{} detected the approximate region, where the rain streaks were located, to guide the separation of the rain layer from the background layer. 

However, early models based on a prior knowledge often suffer from a lack of stability in real scenarios~\cite{kang2011automatic, luo2015removing, li2016rain}. Since 2017, deep learning approaches are developed for rain removal tasks. Deep detail networks~\cite{fu2017removing} narrowed the mapping from input to output and combined prior knowledge to capture high-frequency details, making the model stay focused on rain streaks information. By adding an iterative information feedback network, JORDER~\cite{yang2017deep} used a binary mapping to locate rain streaks. A non-locally enhanced encoder-decoder structure~\cite{li2018non} was proposed to capture long-range dependencies and leverage the hierarchical features of the convolutional layer. In ~\cite{li2018recurrent}, Li \etal{} proposed a deep recurrent convolutional neural network to remove rain marks located at different depths progressively. A density-aware multi-stream connectivity network was introduced for rain removal in~\cite{zhang2018density}. By adding constraints to the cGAN~\cite{isola2017image}, Zhang \etal{}~\cite{zhang2019image} generated more photo-realistic results. A progressive contextual aggregation network~\cite{ren2019progressive} was proposed as a baseline for rain removal. A real-world rain dataset was constructed by Wang \etal{}~\cite{wang2019spatial}, they also incorporated spatial perception mechanisms into deraining networks. Recently, Zhu \etal{}~\cite{zhu2020learning} proposed a gated non-local depth residual network for image rain removal.

\subsection{Single-image Haze Removal}
Similar to image rain removal methods, early work on image dehaze tended to employ statistical methods to acquire prior information by capturing patterns in haze-free images. Representative methods includes Dark channel prior~\cite{he2010single}, color-line prior~\cite{fattal2014dehazing}, colour attenuation prior~\cite{zhu2015fast}, \etc{} However, prior-based methods tend to distort colors and thus produce undesirable artifacts~\cite{he2010single, fattal2014dehazing, zhu2015fast}. In the deep learning era, methods started to not rely on prior knowledge, but to estimate atmospheric light and the transmission map directly. For example, Cai \etal{}~\cite{cai2016dehazenet} proposed an end-to-end dehazing model named DehazeNet, where haze-free images are produced by learning the transmission rate. Similarly, Ren \etal{}~\cite{ren2016single} employed multi-scale deep neural networks to learn the mapping relationship between foggy images and their corresponding transmission maps, aiming to reduce the error in estimating the transmission maps. AODNet~\cite{li2017aod} reconstructed the atmospheric scattering model by leveraging an improved convolutional neural network to learn the mapping relationship between foggy and clean pairs. In~\cite{zhang2018densely}, a single network was proposed to simultaneously learn the intrinsic relationship between transmission maps, atmospheric light, and clean images. Ren \etal{}~\cite{ren2018gated} built an encoder-decoder neural network to enhance the dehazing process. A network with an enhancer and two generators was proposed by Qu \etal{}~\cite{qu2019enhanced}. Chen \etal{}~\cite{chen2019pms} proposed a patch map-based PMS-Net to effectively suppress the color distorted issue.

\subsection{Joint Rain and Haze Removal}

In this line of research, Hu \etal{}~\cite{dgnl2021} built an imaging model for rain streaks and haze based on the visual effect of rain and the scene depth map to synthesize a realistic dataset named RainCityscapes. Han \etal{}~\cite{blind2021han} constructed a superimposed image dataset and proposed a simple yet general Blind Image Decomposition Network to decompose rain streaks, raindrops, and haze in a blind image decomposition setting. Kim \etal{}~\cite{FHRR2021} proposed a frequency-based model for removing rain and haze, where the frequency-based model divided the input image into high-frequency and low-frequency parts with a guided filter and then employed a symmetric encoder-decoder network to remove rain and haze separately. Based on prior knowledge, Liang \etal{}~\cite{chinese2021} proposed a three-stage model, which (1): utilized dark channel prior and depth information to dehaze the low-frequency part of the input image, (2): employed a residual network to remove rain streaks in high-frequency part, and (3): introduced a cGAN~\cite{isola2017image} to refine the local details of the restored image.
\section{Method}

\begin{figure*}[!ht]
\centering
\includegraphics[width=0.98\linewidth]{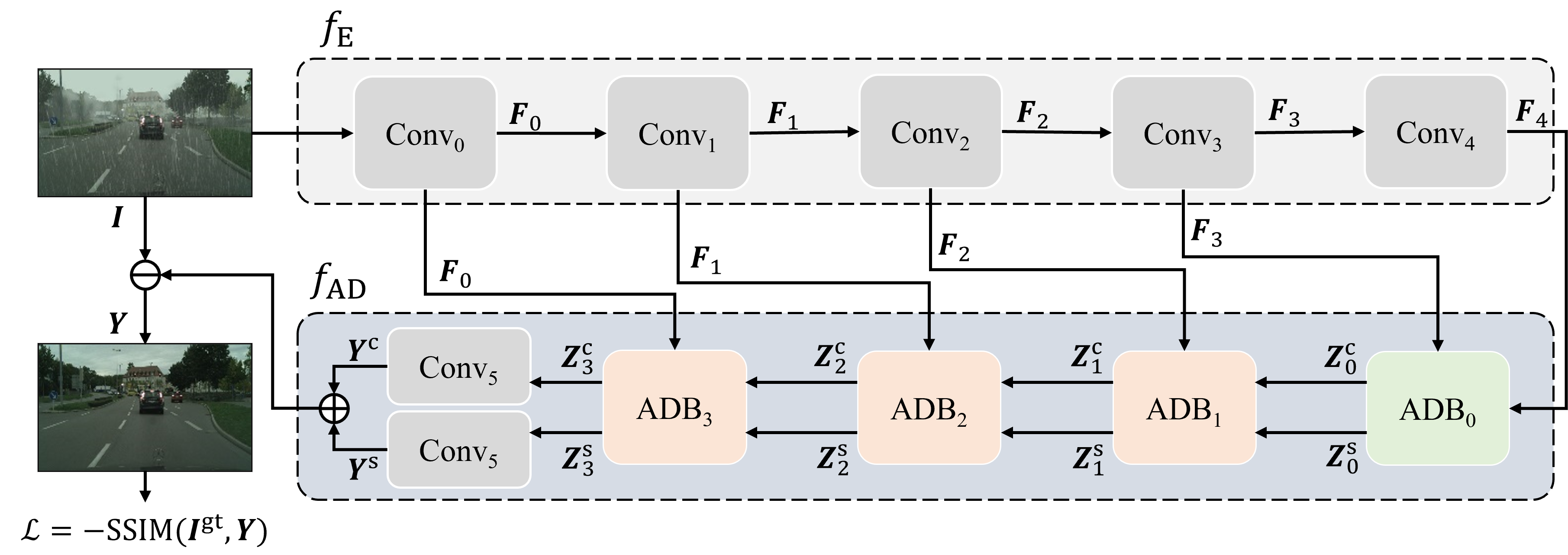}
\caption{The network architecture of the proposed ADU-Net, which consists of a encoder $f_{\mathrm{E}}$ and a asymmetric dual-decoder $f_{\mathrm{AD}}$. $f_{\mathrm{E}}$ has five $\mathrm{Conv}_{i}$ blocks and $f_\mathrm{AD}$ has four $\mathrm{ADB}_j$ blocks and a Conv block. The network is optimized by the SSIM loss function.}
\label{mainfig}
\end{figure*}

This section details the proposed method in a top-down fashion: starting from the problem formulation of our application, followed by the architecture of the proposed \underline{A}symmetric \underline{D}ual-decoder \underline{U-Net} (ADU-Net) and its building block, namely asymmetric dual-decoder block (ADB).

~~\\
\noindent \textbf{Notations.} Throughout the paper, we use bold capital letters to denote matrices or tensors (e.g., $\Mat{X}$), and bold lower-case letters to denote vectors (e.g., $\Vec{x}$).

\subsection{Problem Formulation}

Let a third-order tensor, ${\Mat{I}} \in \mathbb{R}^{C \times H \times W}$, denote an input image, where $C$, $H$ and $W$ present the channel, height, and width of the image, respectively. In our application, both rain and haze are synthesized into the origin scene images as input images. Each input image $\Mat{I}$ is labelled with its ground truth image $\Mat{I}^{\mathrm{gt}}$ without rain and haze in the scene. Our ADU-Net $f_{\theta}$, consisting of a single branch encoder $f_{\mathrm{E}}$, and an asymmetric dual-decoder $f_{\mathrm{AD}}$, can remove the rain and haze in the input image, such that the output of the ADU-Net, $\Mat{Y} = f_{\theta}(\Mat{I})$ can restore its ground truth scene $\Mat{I}^{\mathrm{gt}}$. The ADU-Net is trained to learn a set of parameters, $\theta^*$, with minimum empirical objective value $\mathcal{L}(\Mat{I}^{\mathrm{gt}}, \Mat{Y})$.

\subsection{Network Overview}

We first give a sketch of the proposed ADU-Net. In rain and haze removal applications, one ideal option is to employ the deep neural network to understand the scene of the input image and separate the rain and haze from the input image. In our work, we develop the ADU-Net to remove the rain and haze jointly. As shown in Fig.~\ref{mainfig}, the ADU-Net is stacked by a single branch encoder and an asymmetric dual-decoder. In the encoder $f_{\mathrm{E}}$, we have five convolutional blocks, with each denoted by $\mathrm{Conv}_{i},~0 \leq i \leq 4$.  The output of each convolutional block is denoted by $\Mat{F}_i = \mathrm{Conv}_i(\Mat{F}_{i-1})$ and $\Mat{F}_{-1} = \Mat{I}$.

Then a following asymmetric dual-decoder $f_{\mathrm{AD}}$ aims to recover the scene image without rain and haze (see Fig.~\ref{mainfig}). The proposed asymmetric dual-decoder is stacked of a set of ADBs, which produce two streams of latent features, denoted by $\Mat{Z}^{\mathrm{c}}_j$ and $\Mat{Z}^{\mathrm{s}}_j$ in the $j$-th ADB. Specifically, the processing can be formulated as

\begin{equation}
\Mat{Z}^{\mathrm{c}}_0, \Mat{Z}^{\mathrm{s}}_0 = \mathrm{ADB}_0(\Mat{F}_3, \Mat{F}_4),
\end{equation}
or
\begin{equation}
\Mat{Z}^{\mathrm{c}}_j, \Mat{Z}^{\mathrm{s}}_j = \mathrm{ADB}_j(\Mat{Z}^{\mathrm{c}}_{j-1}, \Mat{Z}^{\mathrm{s}}_{j-1}, \Mat{F}_{3-j}),~j>0.
\end{equation}

After the last ADB, each stream of latent features $\Mat{Z}^{\mathrm{c}}_3$ or $\Mat{Z}^{\mathrm{s}}_3$ is encoded by a convolutional block to recover the channel dimensions into the image space (e.g., $C=3$), as $\Mat{Y}^{\mathrm{c}} = \mathrm{Conv}_5(\Mat{Z}^{\mathrm{c}}_3)$ and $\Mat{Y}^{\mathrm{s}} = \mathrm{Conv}_5(\Mat{Z}^{\mathrm{s}}_3)$. We denote the $\Mat{Y}^{\mathrm{c}}$ as the contamination residual, and $\Mat{Y}^{\mathrm{s}}$ as the scene residual. Having the $\Mat{Y}^{\mathrm{c}}$ and $\Mat{Y}^{\mathrm{s}}$ 
at hand, one can obtain the restored scene image $\Mat{Y}$ as 

\begin{equation}
\Mat{Y} = \Mat{I} - \Mat{Y}^{\mathrm{c}} - \Mat{Y}^{\mathrm{s}}.
\end{equation}

The network is optimized by the negative SSIM loss~\cite{ssim} as $\mathcal{L}_{\mathrm{SSIM}} = -\mathrm{SSIM} (\Mat{I}^{\mathrm{gt}}, \Mat{Y})$. Noted the common practice uses both the negative SSIM loss and MSE loss as the objective. Empirically we observed that a negative SSIM loss works better in the proposed ADU-Net, which will be justified in \textsection~\ref{ss_ablation}.

\subsection{Asymmetric Dual-decoder Block}

In this part, we will describe the asymmetric dual-decoder $f_{\mathrm{AD}}$ in ADU-Net. As shown in Fig.~\ref{mainfig}, $f_{\mathrm{AD}}$ consists of four ADBs and a convolutional block, while the ADBs are stacked by two different instantiations (e.g., $\mathrm{ADB}_0$ vs. $\mathrm{ADB}_j,~j = 1,2,3$). In this following, we will first describe $\mathrm{ADB}_0$, a simple form of the block. Then with a minor modifications, we can realize the $\mathrm{ADB}_j, j = 1,2,3$ on top of the $\mathrm{ADB}_0$.

\begin{figure}[!ht]
\centering
\includegraphics[width=0.8\linewidth]{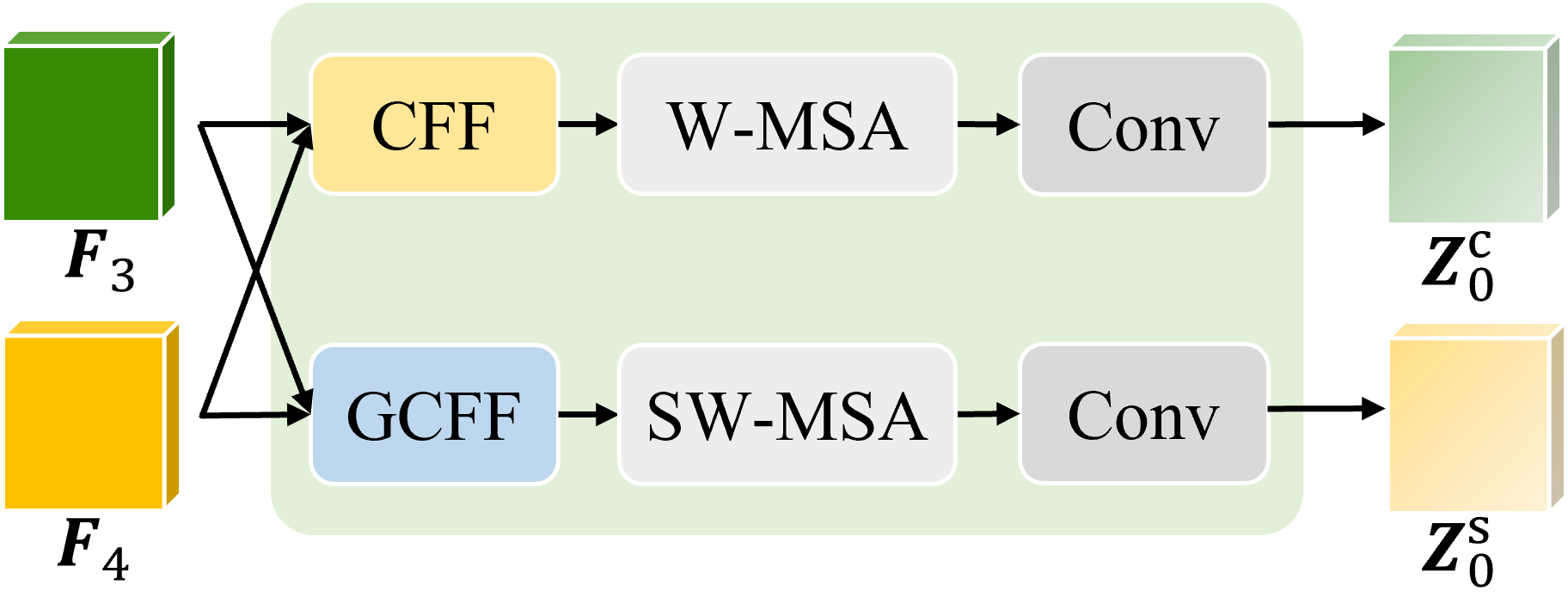}
\caption{Architecture of the first asymmetric dual-block $\mathrm{ADB}_0$. }
\label{dec0}
\end{figure}

The $\mathrm{ADB}_0$ is a two branch architecture (see Fig.~\ref{dec0}), which receives the $\Mat{F}_3$ and $\Mat{F}_4$ as input, and produces two latent features $\Mat{Z}^{\mathrm{c}}_0$ and $\Mat{Z}^{\mathrm{s}}_0$. In $\mathrm{ADB}_0$, the two latent features are respectively encoded by two branch of network, namely contamination residual net (denoted by $g^{\mathrm{c}}$), and scene residual net (denoted by $g^{\mathrm{s}}$), given by
\begin{equation}
\Mat{Z}^{\mathrm{c}}_0 = g^{\mathrm{c}}(\Mat{F}_3, \Mat{F}_4)
\end{equation}
and
\begin{equation}
\Mat{Z}^{\mathrm{s}}_0 = g^{\mathrm{s}}(\Mat{F}_3, \Mat{F}_4).
\end{equation}

\noindent \textbf{Contamination Residual Net.} In the contamination residual net ($g^{\mathrm{c}}$), $\Mat{F}_3$ and $\Mat{F}_4$ are fed to a channel feature fusion (CFF) module to localize the rain and haze areas in the scene image, as
\begin{equation}
\Mat{G}^{\mathrm{c}}_0 = \mathrm{CFF}(\Mat{F}_3, \Mat{F}_4).
\end{equation}

\begin{figure}[!ht]
\centering
\includegraphics[width=0.96\linewidth]{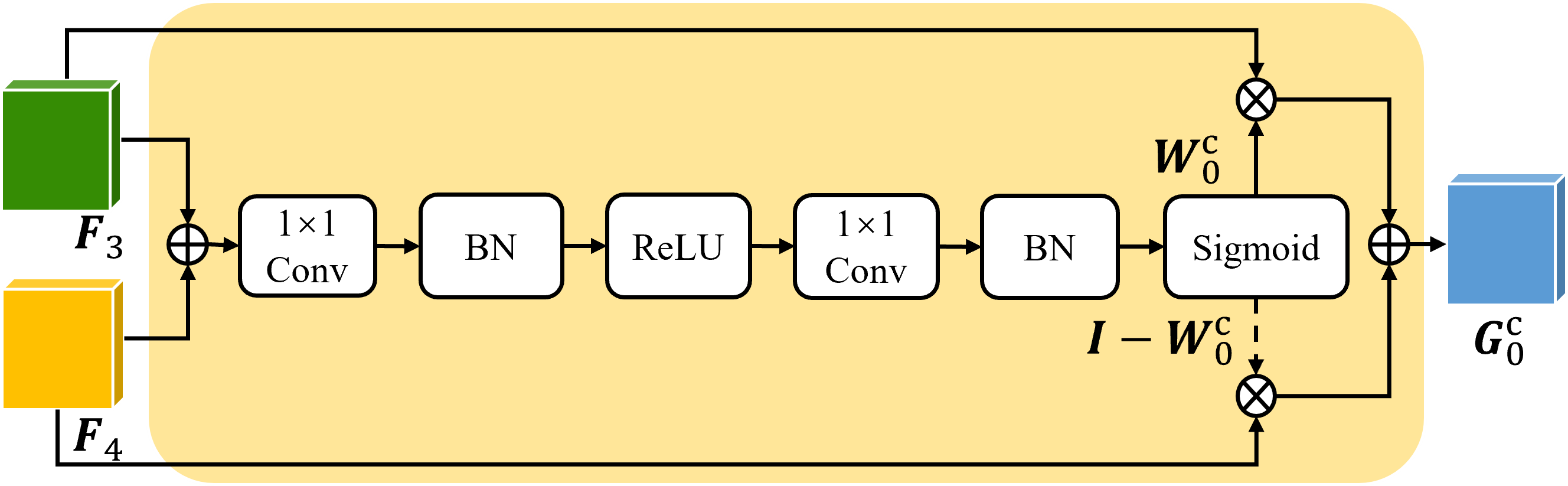}
\caption{Architecture of the channel feature fusion module. }
\label{local}
\end{figure}

The details of CFF are illustrated in Fig.~\ref{local}. Given two feature maps $\Mat{F}_3$ and $\Mat{F}_4$ as input, it first fuses the two inputs by using element-wise addition and then feeds the fused feature maps to 2-layer convolutional blocks to obtain the attention weights, formulated by
\begin{equation}
\Mat{W}_0^{\mathrm{c}} = \sigma\Big(\mathrm{BN}\big(\mathrm{Conv}(\mathrm{ReLU(\mathrm{BN}(\mathrm{Conv}(\Mat{F}_3 \oplus \Mat{F}_4)))})\big)\Big),
\end{equation}
where $\sigma$, $\mathrm{BN}$, $\mathrm{ReLU}$ are sigmoid function, batch normalization, and rectified linear unit activation, respectively. Here, the kernel size of $\mathrm{Conv}$ is $1\times1$, which can be understood as applying a fully-connected layer to the channel features. 

Then we can apply the attention weights to the input feature maps and obtain the fused output, as 
\begin{equation}
\Mat{G}^{\mathrm{c}}_0 = \big( \Mat{W}_0^{\mathrm{c}}  \otimes  \Mat{F}_3 \big) \oplus \big( (\Mat{I} - \Mat{W}_0^{\mathrm{c}}) \otimes \Mat{F}_4 \big).
\end{equation}

The CFF module fuses the input feature maps, and the fusion weights are produced via the channel patterns. We further employ a self-attention mechanism to build the spatially long-range dependencies of the fused feature maps $\Mat{G}^{\mathrm{s}}_0$, given by 

\begin{equation}
\Mat{H}^{\mathrm{c}}_0 = \mathrm{W\text{-}MSA}(\Mat{G}^{\mathrm{c}}_0),
\end{equation}
where W-MSA is the window multi-head self-attention from the Swin Transformer~\cite{swin}.

Having fusing the input feature maps and being processed by the attention mechanism, we can obtain the contamination residual feature maps as 
\begin{equation}
\Mat{Z}^{\mathrm{c}}_0 = \mathrm{Conv}(\Mat{H}^{\mathrm{c}}_0).
\end{equation}

The contamination residual net ($g^{\mathrm{c}}$) aims to attend to the rainy and hazy regions, thereby highlighting the rain and haze components in the contamination residual feature maps.

\noindent \textbf{Scene Residual Net.} Since we can observe from the contamination residual ($\Mat{Y}^{\mathrm{c}}$) that it contains the scene information along with the rain and haze, we develop a scene residual net ($g^{\mathrm{s}}$), that can compensate for the removed scene information in the image. In doing so, the global channel feature fusion (GCFF) module is proposed to capture valuable global scene information of the image, and fuse features, as

\begin{equation}
\Mat{G}^{\mathrm{s}}_0 = \mathrm{GCFF}(\Mat{F}_3, \Mat{F}_4).
\end{equation}

\begin{figure}[!ht]
\centering
\includegraphics[width=0.9\linewidth]{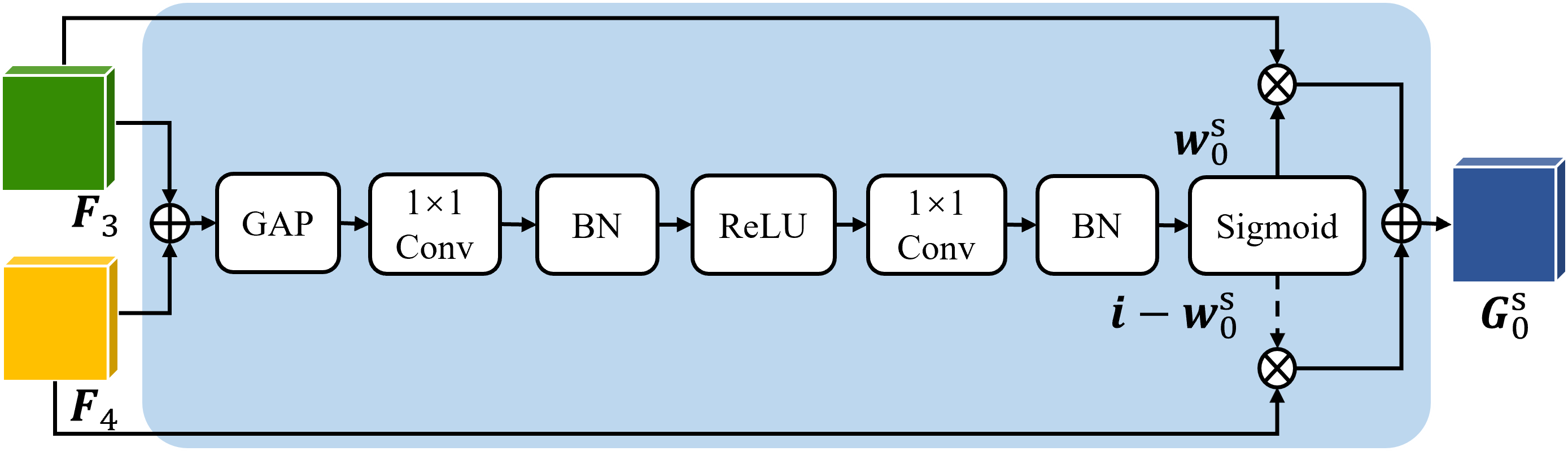}
\caption{Architecture of the global channel feature fusion module. }
\label{global}
\end{figure}

As shown in Fig.~\ref{global}, $\Mat{F}_3$ and $\Mat{F}_4$ are first fused, and summarized to its global feature, as 
\begin{equation}
\Vec{m}^{\mathrm{s}}_0 = \mathrm{GAP}(\Mat{F}_3 \oplus \Mat{F}_4),
\end{equation}
where $\mathrm{GAP}$ indicates the global average pooling. Then a 2-layer convolutional block is used to modulate per element of the global feature $\Vec{m}^{\mathrm{s}}_0$, written by:
\begin{equation}
\Vec{w}^{\mathrm{s}}_0 = \sigma\Big(\mathrm{BN}\big(\mathrm{Conv}(\mathrm{ReLU(\mathrm{BN}(\mathrm{Conv}(\Vec{m}^{\mathrm{s}}_0)))})\big)\Big).
\end{equation}

We can thereby fuse the input feature maps as:
\begin{equation}
\Mat{G}^{\mathrm{s}}_0 = \big( \Vec{w}^{\mathrm{s}}_0  \otimes  \Mat{F}_3 \big) \oplus \big( (\Vec{i} - \Vec{w}_0^{\mathrm{s}}) \otimes \Mat{F}_4 \big).
\end{equation}

In GCFF, we employ the shift-window multi-head self-attention (SW-MSA) to  enhance the spatial interaction of the feature maps and obtain the scene residual features, described by

\begin{equation}
\Mat{H}^{\mathrm{s}}_0 = \mathrm{SW\text{-}MSA}(\Mat{G}^{\mathrm{s}}_0),
\end{equation}
and 
\begin{equation}
\Mat{Z}^{\mathrm{s}}_0 = \mathrm{Conv}(\Mat{H}^{\mathrm{s}}_0).
\end{equation}

\begin{figure}[!ht]
\centering
\includegraphics[width=0.98\linewidth]{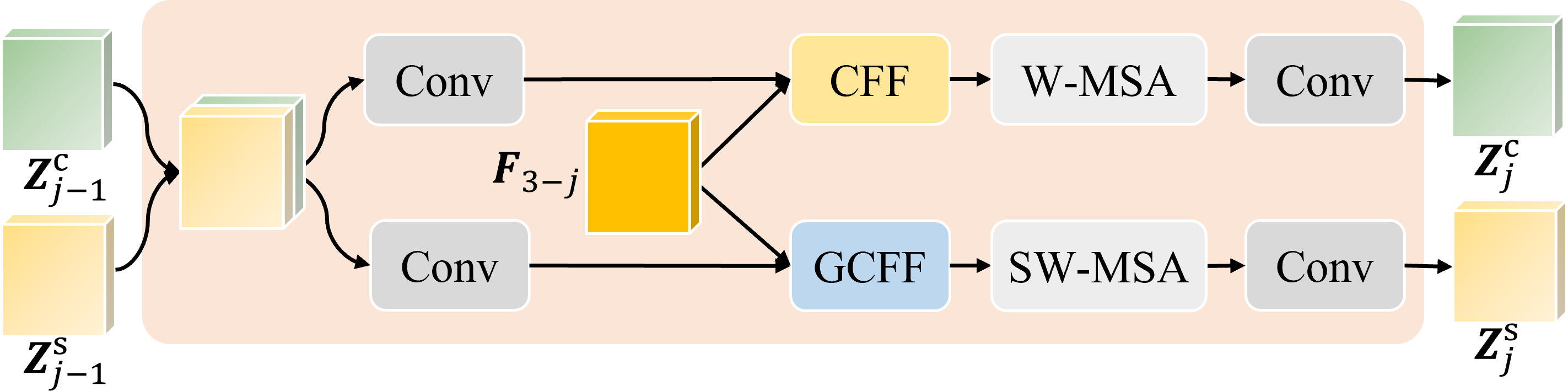}
\caption{Architecture of asymmetric dual-block $\mathrm{ADB}_j$.}
\label{deci}
\end{figure}

\noindent \textbf{Instantiation of $\Vec{ADB}_j$.} The difference between $\mathrm{ADB}_{j}, j \neq 0$ and $\mathrm{ADB}_0$ is that $\mathrm{ADB}_0$ receives two feature maps as input, while $\mathrm{ADB}_j, j\neq 0$ includes three feature maps as input. To adapt the architecture of $\mathrm{ADB}_0$ to $\mathrm{ADB}_j, j \neq 0$, we make minor modification (see Fig.~\ref{deci}). Specifically, for any block, $\mathrm{ADB}_j$, its input includes the output from $j-1$-th ADB blcok, e.g., $\Mat{Z}^{\mathrm{c}}_{j-1}, \Mat{Z}^{\mathrm{s}}_{j-1} \in \mathbb{R}^{d \times h\times w}$, and from the $3-j$-th convolutional encoder, e.g., $\Mat{F}_{3-j}$. We first concatenate the $\Mat{Z}^{\mathrm{c}}_{j-1}, \Mat{Z}^{\mathrm{s}}_{j-1}$, and reduce its dimension from $2d \times h \times w$ to $d \times h \times w$, as
\begin{equation}
\bar{\Mat{Z}}_{j-1} = \mathrm{Concat}(\Mat{Z}^{\mathrm{c}}_{j-1}, \Mat{Z}^{\mathrm{s}}_{j-1}).
\end{equation}
and
\begin{equation}
\tilde{\Mat{Z}}^{\mathrm{c}}_{j-1} = \mathrm{Conv_{in}}(\bar{\Mat{Z}}_{j-1}),~\tilde{\Mat{Z}}^{\mathrm{s}}_{j-1} = \mathrm{Conv_{in}}(\bar{\Mat{Z}}_{j-1}).{}\end{equation}
With $\Mat{F}_{3-j}$, the output of $\mathrm{ADB}_j$ can be obtained as
\begin{equation}
\begin{split}
\Mat{Z}^{\mathrm{c}}_j &= g^{\mathrm{c}}(\tilde{\Mat{Z}}^{\mathrm{c}}_{j-1} ,\Mat{F}_{3-j}) \\
&= \mathrm{Conv_{out}}\Big(\mathrm{W\text{-}MSA}\big(\mathrm{CFF}(\tilde{\Mat{Z}}^{\mathrm{c}}_{j-1} ,\Mat{F}_{3-j})\big)\Big)
\end{split}
\end{equation}
and
\begin{equation}
\begin{split}
\Mat{Z}^{\mathrm{s}}_j &= g^{\mathrm{s}}(\tilde{\Mat{Z}}^{\mathrm{s}}_{j-1} ,\Mat{F}_{3-j}) \\
&= \mathrm{Conv_{out}}\Big(\mathrm{SW\text{-}MSA}\big(\mathrm{GCFF}(\tilde{\Mat{Z}}^{\mathrm{s}}_{j-1} ,\Mat{F}_{3-j})\big)\Big).
\end{split}
\end{equation}

In this work, we propose a novel architecture for the rain and haze removal task. Considering the network capacity and hardware overhead, we propose two sizes of networks. One is the lite network, called ADU-Net, and another one is the large network, called ADU-Net-plus. In \textsection~\ref{s_expt}, we present the details of two architectures. The network performance is also evaluated in \textsection~\ref{s_expt}.

\begin{figure*}[ht]
\centering
\subfloat[Residual U-Net~\cite{chen2021robust}]{\includegraphics[width=.25\linewidth]{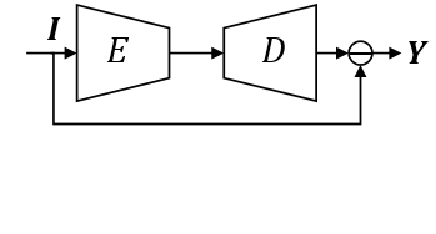}\label{fig.1}}%
\quad
\subfloat[Dual-decoder U-Net]{\includegraphics[width=.3\linewidth]{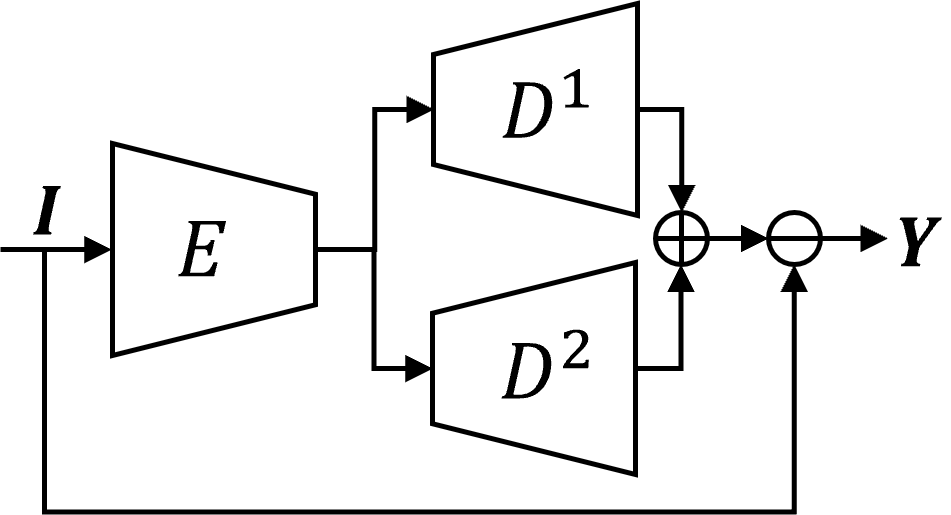}\label{fig.2}}%
\quad
\subfloat[ADU-Net]{\includegraphics[width=.3\linewidth]{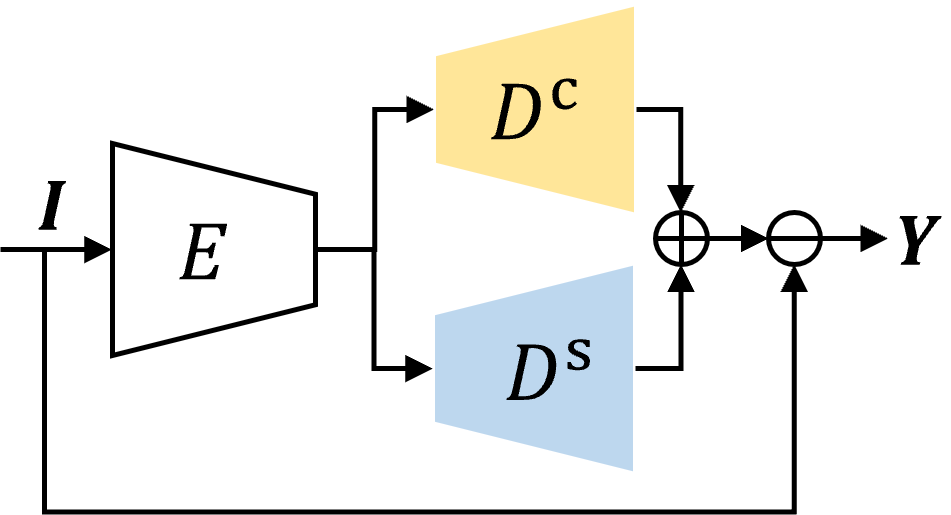}\label{fig.3}}
\caption{Schematic comparison of the ADU-Net architecture and U-Net-based architectures. (a) is a vanilla architecture of the residual U-Net. (b) is a simple form of the residual U-Net with dual decoders. (c) is the diagram of our method.}\label{fig:diff}
\end{figure*}

\begin{remark}
The residual U-Net architecture has been used extensively for the rain or haze removal tasks~\cite{chen2021robust}, as shown in~Fig.~\ref{fig.1}. Having the observation that the contamination residual, produced by the decoder, contains the scene information, we aim to develop a dual-decoder U-Net, with one decoder producing the contamination residual, and another one producing the scene residual as a scene compensator. Its initial design is shown in Fig.~\ref{fig.2}. Considering the physical property of the contamination and scene information in the input image, we propose a novel network architecture, ADU-Net, where we integrate two decoders with non-identical architectures (see Fig.~\ref{fig.3}). We justify our design in \textsection~\ref{ss_ablation}.
\end{remark}

\sloppy
\section{Experiments}\label{s_expt}

In this section, we first give the implementation details of the proposed ADU-Net and ADU-Net-plus. Then the benchmark datasets and evaluation protocol are also introduced. We further compare our network to the state-of-the-art methods and conduct ablation studies to evaluate the superiority of the proposed network and each components. In the final part, we demonstrate substantial qualitative results to analyze the superior performance of our network.

\subsection{Implementation Details}\label{ss_implementation}
\textbf{Network Architecture. } 
The overall neural architecture of the proposed network is shown in Fig.~\ref{mainfig}. Table~\ref{net} lists the kernel size of the convolutional layers. In the encoder block, the feature maps are processed by the Batch Normalization~\cite{BatchNA} and ReLU~\cite{relu} after the convolutional layer, i.e., $\mathrm{Conv_0}$, $\mathrm{Conv_1}$, $\mathrm{Conv_2}$, $\mathrm{Conv_3}$, $\mathrm{Conv_4}$. Then the max-pooling layer is employed to down-sample the feature maps in each layer. In the decoder block, we also list the kernel size in the convolutional layers (see Table~\ref{net}), and employ the Leaky ReLU as the activation function. Having the computational efficiency in mind, we develop two neural networks of different scales. The light one is denoted as ADU-Net, while the large one is denoted as ADU-Net-plus. As shown in Table~\ref{net}, the difference between the two networks is merely the modification to the channel dimensions. The superiority of our network will be evaluated in \textsection~\ref{sec:sota}.

\begin{table}[!t]
\caption{Details of the kernel size in convolution layers. $H$ and $W$ denote the height and width of the input image, respectively. }\label{net}
\centering
\scalebox{0.95}{
\begin{tabular}{c|c|c|c|c}
\hline
\multicolumn{2}{c|}{Layer name} & Output size & ADU-Net  & ADU-Net-plus \\ \hline
\multicolumn{2}{c|}{$\mathrm{Conv_0}$}      & $H \times W$         & $\left[\begin{array}{l}3 \times 3,32 \\ 3 \times 3,32\end{array}\right]$ & $\left[\begin{array}{l}3 \times 3,64 \\ 3 \times 3,64\end{array}\right]$           \\ \hline
\multicolumn{2}{c|}{$\mathrm{Conv_1}$}     & $\frac{H}{2} \times \frac{W}{2}$ & $\left[\begin{array}{l}3 \times 3,64 \\ 3 \times 3,64\end{array}\right]$ & $\left[\begin{array}{l}3 \times 3,128 \\ 3 \times 3,128\end{array}\right]$           \\ \hline
\multicolumn{2}{c|}{$\mathrm{Conv_2}$}     & $\frac{H}{4} \times \frac{W}{4}$ & $\left[\begin{array}{l}3 \times 3,128 \\ 3 \times 3,128\end{array}\right]$ & $\left[\begin{array}{l}3 \times 3,256 \\ 3 \times 3,256\end{array}\right]$           \\ \hline
\multicolumn{2}{c|}{$\mathrm{Conv_3}$}     & $\frac{H}{8} \times \frac{W}{8}$ & $\left[\begin{array}{l}3 \times 3,256 \\ 3 \times 3,256\end{array}\right]$ & $\left[\begin{array}{l}3 \times 3,512 \\ 3 \times 3,512\end{array}\right]$           \\ \hline
\multicolumn{2}{c|}{$\mathrm{Conv_4}$}      & $\frac{H}{16} \times \frac{W}{16}$ & $\left[\begin{array}{l}3 \times 3,256 \\ 3 \times 3,256\end{array}\right]$ & $\left[\begin{array}{l}3 \times 3,512 \\ 3 \times 3,512\end{array}\right]$           \\ \hline
\multicolumn{2}{c|}{$\mathrm{ADB_0}$}       & $\frac{H}{8} \times \frac{W}{8}$ & $\left[\begin{array}{l}3 \times 3,128 \\ 3 \times 3,128\end{array}\right]$ & $\left[\begin{array}{l}3 \times 3,256 \\ 3 \times 3,256\end{array}\right]$           \\ \hline
\multirow{2}{*}{$\mathrm{ADB_1}$} 
& $\mathrm{Conv_{in}}$       & $\frac{H}{4} \times \frac{W}{4}$ & $\left[\begin{array}{l}3 \times 3,128 \\ 3 \times 3,128\end{array}\right]$ & $\left[\begin{array}{l}3 \times 3,256 \\ 3 \times 3,256\end{array}\right]$           \\ \cline{2-5}
& $\mathrm{Conv_{out}}$       & $\frac{H}{4} \times \frac{W}{4}$ & $\left[\begin{array}{l}3 \times 3,64 \\ 3 \times 3,64\end{array}\right]$ & $\left[\begin{array}{l}3 \times 3,128 \\ 3 \times 3,128\end{array}\right]$           \\ \hline
\multirow{2}{*}{$\mathrm{ADB_2}$} 
& $\mathrm{Conv_{in}}$      & $\frac{H}{2} \times \frac{W}{2}$ & $\left[\begin{array}{l}3 \times 3,64 \\ 3 \times 3,64\end{array}\right]$ & $\left[\begin{array}{l}3 \times 3,128 \\ 3 \times 3,128\end{array}\right]$           \\ \cline{2-5}
& $\mathrm{Conv_{out}}$       & $\frac{H}{2} \times \frac{W}{2}$ & $\left[\begin{array}{l}3 \times 3,32 \\ 3 \times 3,32\end{array}\right]$ & $\left[\begin{array}{l}3 \times 3,64 \\ 3 \times 3,64\end{array}\right]$           \\ \hline
\multirow{2}{*}{$\mathrm{ADB_3}$} 
& $\mathrm{Conv_{in}}$       & $H \times W$         & $\left[\begin{array}{l}3 \times 3,32 \\ 3 \times 3,32\end{array}\right]$ & $\left[\begin{array}{l}3 \times 3,64 \\ 3 \times 3,64\end{array}\right]$           \\ \cline{2-5}
& $\mathrm{Conv_{out}}$       & $H \times W$         & $\left[\begin{array}{l}3 \times 3,16 \\ 3 \times 3,16\end{array}\right]$ & $\left[\begin{array}{l}3 \times 3,32 \\ 3 \times 3,32\end{array}\right]$           \\ \hline
\multicolumn{2}{c|}{$\mathrm{Conv_5}$}      & $H \times W$         & $\left[\begin{array}{l}3 \times 3,3 \\ 3 \times 3,3\end{array}\right]$ & $\left[\begin{array}{l}3 \times 3,3 \\ 3 \times 3,3\end{array}\right]$           \\ \hline
\multicolumn{3}{c|}{Parameter size} & $6.63\times 10^6$ & $26.45 \times 10^6$\\ \hline
\end{tabular}
}
\end{table}

\textbf{Network Training. } We implement our method using PyTorch deep learning package~\cite{pytorch}. All experiments are evaluated on NVIDIA GTX 2080ti GPUs. In the experiments for RainCityscapes~\cite{dgnl2021} and BID Rain datasets~\cite{blind2021han}, the input images are resized to $512 \times 256$. For the SPA-Data, we follow the practice in~\cite{wang2019spatial}, that uses original images with size of $256 \times 256$. 
The Adam optimization scheme with an initial learning rate of 0.001 is used to optimize the network. We train the network for 100 epochs for RainCityscapes and BID Rain datasets, and 20 epochs for SPA-Data. The learning rate adjustment strategy is employed to realize the learning rate decay, where the learning rate is decayed by a factor of 0.1 when the accuracy of the network does not improve in 5 epochs.

\begin{table*}[!ht]
\centering
\caption{The Statistics of Datasets. }\label{tab:data}
\scalebox{1.0}{
\begin{tabular}{c|c|c|c|c|cccc}
\hline
\multirow{2}{*}{Dataset} & \multirow{2}{*}{\begin{tabular}[c]{@{}c@{}}Train set\end{tabular}} & \multirow{2}{*}{\begin{tabular}[c]{@{}c@{}}Test set\end{tabular}} & \multicolumn{2}{c|}{Property}               & \multicolumn{4}{c}{Contamination}                                                                                                                          \\ \cline{4-9} 
                         &                                                                      &                                                                     & \multicolumn{1}{c|}{Synthetic} & Real world & \multicolumn{1}{c|}{\begin{tabular}[c]{@{}c@{}}Rain  streaks\end{tabular}} & \multicolumn{1}{c|}{Haze}      & \multicolumn{1}{c|}{Snow}      & Raindrops \\ \cline{1-9}
RainCityscapes          & 9,432                                                                & 1,188                                                               & \multicolumn{1}{c|}{\checkmark} &            & \multicolumn{1}{c|}{\checkmark}                                               & \multicolumn{1}{c|}{\checkmark} & \multicolumn{1}{c|}{}          &           \\ \cline{1-9}
BID Rain                 & 2,975                                                                & 500 * 6                                                                & \multicolumn{1}{c|}{\checkmark} &            & \multicolumn{1}{c|}{\checkmark}                                               & \multicolumn{1}{c|}{\checkmark} & \multicolumn{1}{c|}{\checkmark} & \checkmark \\ \cline{1-9}
SPA-Data                 & 638,492                                                              & 1,000                                                               & \multicolumn{1}{c|}{}          & \checkmark  & \multicolumn{1}{c|}{\checkmark}                                               & \multicolumn{1}{c|}{}          & \multicolumn{1}{c|}{}          &           \\ \cline{1-9}
\end{tabular}
}
\end{table*}

\subsection{Datasets and Evaluation Protocol}
We evaluate the proposed methods on two synthetic datasets, i.e., RainCityscapes~\cite{dgnl2021}, BID Rain~\cite{blind2021han}, and a real-world dataset SPA-Data~\cite{wang2019spatial}. In the following, we will introduce these datasets and the statistics of each dataset are illustrated in Table~\ref{tab:data}.

\textbf{RainCityscapes.} 
The RainCityscapes dataset is synthesized from the Cityscapes dataset~\cite{cityscape}. It takes 9,432 images synthesized from 262 Cityscapes images as the training set and 1,188 images synthesized from 33 Cityscapes images as the test set. All the selected images of Cityscapes are overcast, without obvious shadow. Rain streaks and haze is synthesized by different intensity maps. By adjusting the intensity of the rain streaks and haze, each original image can produce 36 different synthesized images. The results of different methods are reported in Table~\ref{SA}.

\textbf{BID Rain.} 
The BID Rain dataset is also synthesized from the Cityscapes dataset. It samples 2,975 images from the validation set of the Cityscapes dataset as a training set, and 500 images from the test set of the Cityscapes dataset as its test set. This is a complicated dataset as the images contain rain streaks, haze, snow, and raindrops. The rain streaks masks are sampled from Rain100L and Rain100H~\cite{yang2017deep}, and the snow masks are sampled from Snow 100K~\cite{snow100}. The haze masks include three different intensities originating from FoggyCityScape~\cite{foggycityscape}. The raindrops are produced from the metaball model~\cite{metaball}. Those weather components are mixed with the images in the Cityscapes dataset using the physical imaging models~\cite{yang2017deep, snow100, foggycityscape, he2010single, metaball}. In the training set, every image can be mixed with each weather component with random probabilities, and we evaluate our model in six different cases, the combinations of the weather components in each case are as follows (1): rain streaks, (2): rain streaks and snow, (3): rain streaks and light haze, (4): rain streaks and heavy haze, (5): rain streaks, moderate haze and raindrops and (6): rain streaks, snow, moderate haze and raindrops. Refer~\cite{blind2021han} for more details of six settings. The results of different cases are shown in Table~\ref{bidset}.

\textbf{SPA-Data.} 
The SPA-Data is a real-world dataset, which is cropped from 170 real rain videos, of which 86 videos are collected from StoryBlocks or YouTube, and 84 videos are captured by iPhone X or iPhone 6SP. Those videos cover outdoor fields, suburb scenes, and common urban scenes. This dataset contains 638,492 image pairs for training and 1,000 for testing. The results of SPA-Data in Table~\ref{spa}.

In our experiments, the network performance is quantitatively evaluated by the peak signal-to-noise ratio (PSNR) and structural similarity (SSIM) metrics. A higher value of PSNR and SSIM indicates a better image recovery performance of the network.

\subsection{Comparison to the State-of-the-Arts}\label{sec:sota}
To verify the advance of our method, we compare the performance of our method with current state-of-the-art methods across three datasets. 

\textbf{RainCityscapes. }
In the RainCityscapes dataset, we compare our methods to the the state-of-the-art rain removal methods including RESCA~\cite{li2018non}, PReNet~\cite{ren2019progressive}, DuRN~\cite{liu2019dual}, RCDNet~\cite{rcdnet2020wang}, SPANet~\cite{wang2019spatial} and MPRNet~\cite{zamir2021multi}. We also compare our methods with approaches that jointly remove the rain and haze, i.e., DAF-Net~\cite{dafhu2019depth}, DGNL-Net~\cite{dgnl2021}. The comparison with haze removal methods, like EPDN~\cite{qu2019enhanced}, DCPDN~\cite{zhang2018densely}, AECR-Net~\cite{AECRNET}, is also conducted. The results are reported in Table~\ref{SA}. We can find that our vanilla solution, i.e. ADU-Net, outperforms the existing state-of-the-art methods. In particular, it improves the PSNR/SSIM values of the DGNL-Net by 1.45/0.0041, indicating the superior design of our method. The plus version of our method, i.e., ADU-Net-plus, again brings performance gain over the AUD-Net, where the ADU-Net-plus improves the PSNR/SSIM values by 0.81/0.0021.

\begin{table}[!ht]
\caption{Comparison with the State-of-the-Arts Methods of rain removal and haze removal on RainCityscapes dataset. $^\dag$ indicates the network was trained on the RainCityscapes dataset. $^\ddag$ indicates the results of the algorithms as reported in~\cite{dgnl2021} $1^{\mathrm{st}}/2^{\mathrm{nd}}$ best in \textcolor{red}{red}/\textcolor{blue}{blue}.}\label{SA}
\centering
\begin{tabular}{c |c| c| c }
\hline
\multicolumn{2}{c|}{Method}  & PSNR & SSIM \\ \hline
\multicolumn{2}{c|}{Input}  & 15.55 & 0.7722 \\ 
 \hline
 \multirow{3}{*}{Haze removal}
 &EPDN$^{\ddag}$\cite{qu2019enhanced} & 26.08 &0.9306 \\ \cline{2-4}
 &DCPDN$^{\ddag}$\cite{zhang2018densely} &28.52 &0.9277 \\ \cline{2-4}
 &AECRNet$^{\dag}$\cite{AECRNET} &28.77 &0.9350 \\
 \hline
\multirow{6}{*}{Rain removal}
 &RESCAN${^{\ddag}}$\cite{li2018recurrent} & 24.49& 0.8852 \\ \cline{2-4}
 &PReNet${^{\dag}}$\cite{ren2019progressive} & 27.34& 0.9497\\ \cline{2-4}
 &DuRN${^{\ddag}}$\cite{liu2019dual} & 29.43 & 0.9487\\ \cline{2-4}
 &RCDNet${^{\dag}}$\cite{rcdnet2020wang}&30.56 &0.8873 \\ \cline{2-4}
 &SPANet${^{\ddag}}$\cite{wang2019spatial} & 31.48&0.9656\\ \cline{2-4}
 &MPRNet${^{\dag}}$\cite{zamir2021multi}&32.33 &0.9767 \\ 
 \hline
 \multirow{4}{*}{Rain and haze removal}
 &DAF-Net${^{\dag}}$\cite{dafhu2019depth} & 30.16& 0.9531\\ \cline{2-4}
 &DGNL-Net${^{\dag}}$\cite{dgnl2021} & 32.38& 0.9743\\  \cline{2-4}
 &ADU-Net & \textcolor{blue}{33.83} &  \textcolor{blue}{0.9784}\\ \cline{2-4}
 &ADU-Net-plus & \textcolor{red}{34.64} &  \textcolor{red}{0.9805}\\ 
 \hline 
\end{tabular}
\end{table}

\textbf{BID Rain. } Since the scene in the RainCityscapes dataset only contains rain and haze information, we further evaluate our methods on the challenging dataset, BID Rain, to verify its generalization of working in complicated weather conditions. Table~\ref{bidset} illustrates the comparison of the model performance in each weather condition. We can observe that the proposed ADU-Net can outperform the BIDeN~\cite{blind2021han} in each of the cases. Especially in cases (2) and (3), the ADU-Net brings the maximum performance gain. One possible explanation is that the proposed ADU-Net is designed with dual-decoder, which is tailored for the images in case (2) including the rain streaks and snow, or that in (3) including rain streaks and a light haze. However, the improvement in the other cases reveals the generalization of our proposal. Along with the ADU-Net, its plus version can significantly improve both PSNR/SSIM values, showing the superiority of our network architecture. In case (4), the performance of ADU-Net is lower than that of BIDeN. One possible explanation is that the ``heavy haze'' covers the scenes, which makes it difficult for our network to produce the scene residual. Nevertheless, this issue is addressed by increasing the parameter size, supported by the performance in ADU-Net-plus.

\begin{table*}[!t]
\centering
\caption{Comparison with the State-of-the-Arts Methods on BID Rain dataset. $^\dag$ indicates the network was trained on the BID Rain dataset. $1^{\mathrm{st}}/2^{\mathrm{nd}}$ best in \textcolor{red}{red}/\textcolor{blue}{blue}.}\label{bidset}
\begin{tabular}{c|cc|cc|cc|cc|cc|cc}
\hline
\multirow{2}{*}{Case} & \multicolumn{2}{c|}{Input}          & \multicolumn{2}{c|}{PReNet${^{\dag}}$\cite{ren2019progressive}}         & \multicolumn{2}{c|}{RCDNet${^{\dag}}$\cite{rcdnet2020wang}}         & \multicolumn{2}{c|}{BIDeN${^{\dag}}$\cite{blind2021han}}          & \multicolumn{2}{c|}{ADU-Net}        & \multicolumn{2}{c}{ADU-Net-plus}                      \\ \cline{2-13} 
                      & \multicolumn{1}{c|}{PSNR}  & SSIM   & \multicolumn{1}{c|}{PSNR}  & SSIM   & \multicolumn{1}{c|}{PSNR}  & SSIM   & \multicolumn{1}{c|}{PSNR}  & SSIM   & \multicolumn{1}{c|}{PSNR}  & SSIM   & \multicolumn{1}{c|}{PSNR}           & SSIM            \\ \hline
(1)                   & \multicolumn{1}{c|}{25.51} & 0.8144 & \multicolumn{1}{c|}{32.69} & 0.9803 & \multicolumn{1}{c|}{28.05} & 0.9527 & \multicolumn{1}{c|}{31.17} & 0.9438 & \multicolumn{1}{c|}{\textcolor{blue}{34.62}} & \textcolor{blue}{0.9827} & \multicolumn{1}{c|}{\textcolor{red}{39.05}} & \textcolor{red}{0.9877} \\ \hline
(2)                   & \multicolumn{1}{c|}{18.69} & 0.5979 & \multicolumn{1}{c|}{30.52} & 0.9504 & \multicolumn{1}{c|}{29.84} & 0.9351 & \multicolumn{1}{c|}{29.47} & 0.9089 & \multicolumn{1}{c|}{\textcolor{blue}{32.47}} & \textcolor{blue}{0.9560} & \multicolumn{1}{c|}{\textcolor{red}{36.48}} & \textcolor{red}{0.9742} \\ \hline
(3)                   & \multicolumn{1}{c|}{17.48} & 0.7427 & \multicolumn{1}{c|}{29.65} & 0.9568 & \multicolumn{1}{c|}{30.17} & 0.9536 & \multicolumn{1}{c|}{28.90} & 0.9325 & \multicolumn{1}{c|}{\textcolor{blue}{31.48}} & \textcolor{blue}{0.9669} & \multicolumn{1}{c|}{\textcolor{red}{33.75}} & \textcolor{red}{0.9777} \\ \hline
(4)                   & \multicolumn{1}{c|}{11.55} & 0.6017 & \multicolumn{1}{c|}{25.80} & 0.9233 & \multicolumn{1}{c|}{{26.74}} & 0.9210  & \multicolumn{1}{c|}{\textcolor{blue}{26.82}} & 0.9125 & \multicolumn{1}{c|}{{26.52}} & \textcolor{blue}{0.9360} & \multicolumn{1}{c|}{\textcolor{red}{29.30}} & \textcolor{red}{0.9565} \\ \hline
(5)                   & \multicolumn{1}{c|}{14.02} & 0.6455 & \multicolumn{1}{c|}{27.36} & 0.9302 & \multicolumn{1}{c|}{28.30} & 0.9285 & \multicolumn{1}{c|}{27.31} & 0.9116 & \multicolumn{1}{c|}{\textcolor{blue}{28.54}} & \textcolor{blue}{0.9443} & \multicolumn{1}{c|}{\textcolor{red}{30.32}} & \textcolor{red}{0.9594} \\ \hline
(6)                   & \multicolumn{1}{c|}{12.38} & 0.4916 & \multicolumn{1}{c|}{26.56} & 0.9046 & \multicolumn{1}{c|}{27.26} & 0.9005 & \multicolumn{1}{c|}{26.54} & 0.8675 & \multicolumn{1}{c|}{\textcolor{blue}{27.63}} & \textcolor{blue}{0.9222} & \multicolumn{1}{c|}{\textcolor{red}{29.66}} & \textcolor{red}{0.9418} \\ \hline
\end{tabular}
\end{table*}

\textbf{SPA-Data. } 
We also evaluate our methods in the large-scale dataset, SPA-Data. We compare our methods to the existing state-of-the-art methods in Table~\ref{spa}, including RESCAN\cite{li2018recurrent}, PReNet\cite{ren2019progressive}, SPANet\cite{wang2019spatial} and RCDNet\cite{rcdnet2020wang}. As shown in Table~\ref{spa}, the proposed methods outperform the existing methods by a large margin. For example, the improvements read of 2.72/0.0051  (PSNR/SSIM) from ADU-Net and 4.57/0.0090 from ADU-Net-plus, as compared to RCDNet, showing the strong performance of our network architecture.

\begin{table}
\centering
\caption{Comparison with the State-of-the-Arts Methods on SPA-Data dataset. $^\ddag$ indicates the results of the algorithms as reported in~\cite{rcdnet2020wang}. $1^{\mathrm{st}}/2^{\mathrm{nd}}$ best in \textcolor{red}{red}/\textcolor{blue}{blue}.}\label{spa}
\begin{tabular}{ c c c}
\hline
 Method & PSNR & SSIM \\ 
 \hline
 Input & 34.15 & 0.9269\\
 \hline
 RESCA$\mathrm{N}^\ddag$\cite{li2018recurrent}& 38.19 & 0.9707 \\
 \hline
 PReNe$\mathrm{t}^\ddag$\cite{ren2019progressive} & 40.16 & 0.9816\\
 \hline
 SPANe$\mathrm{t}^\ddag$\cite{wang2019spatial} & 40.24 &0.9811\\
 \hline 
 RCDNe$\mathrm{t}^\ddag$\cite{rcdnet2020wang} & 41.47 & 0.9834\\
 \hline
 ADU-Net& \textcolor{blue}{44.19} & \textcolor{blue}{0.9885}\\
 \hline
 ADU-Net-plus & \textcolor{red}{46.04} & \textcolor{red}{0.9924}\\
 \hline   
\end{tabular}
\end{table}

\subsection{Ablation Study}\label{ss_ablation}

In this section, we conduct thorough ablation studies to verify the effectiveness per component in the proposed network. All studies in this section are conducted using ADU-Net on the RainCityscapes dataset.

\textbf{Loss Function.} 
In our implementation, the network is optimized by the negative SSIM loss, i.e., $\mathcal{L}_{\mathrm{SSIM}}$. While in many practices of the low-level computer vision tasks, the MSE loss i.e., $\mathcal{L}_{\mathrm{MSE}}$, is employed~\cite{ssim+mse}. In this study, we evaluate the effectiveness of each loss function. As shown in Table~\ref{SSIM-MSE}, we can find that each of the loss functions works better for our rain and haze removal task, and the network performance training from the two-loss functions are similar. However, the multi-task training, which optimizes the loss functions jointly, will degrade the network performance, indicating that the network may be saturated using one loss function, and the joint training will harm the network.

\begin{table}[!t]
\caption{Comparison of the effectiveness of Loss Functions. We use \textbf{bold} to indicate best the result. }
\centering\label{SSIM-MSE}
\begin{tabular}{l c c c}
\hline
 Loss Function & $\mathcal{L}_{\mathrm{MSE}}$ & $\mathcal{L}_{\mathrm{SSIM}}$ & $\mathcal{L}_{\mathrm{MSE}} + \mathcal{L}_{\mathrm{SSIM}}$\\ 
 \hline
 PSNR & 33.17 & \textbf{33.83} &33.74\\ 
 \hline 
 SSIM & 0.9720 & \textbf{0.9784} &0.9774\\
 \hline   
\end{tabular}
\end{table}

\textbf{Effect of Dual-branch Architecture. } Our work naively proposes a dual-branch architecture, i.e., asymmetric dual-decoder U-Net, for rain and haze removal tasks. In this study, we will justify the effectiveness of the dual-branch design in our task (shown in Fig~\ref{fig:diff}). Table~\ref{dual_1} shows the empirical comparison of three architectures, i.e., Residual U-Net, Dual-decoder U-Net, and the proposed ADU-Net. Table~\ref{dual_1} verifies our design is reasonable, where the dual-decoder U-Net outperforms the vanilla version of the residual U-Net and our ADU-Net can further bring the performance gain to the dual-decoder U-Net. 

\begin{table}[!ht]
\caption{Effect of dual-branch architecture in rain and haze removal. We use \textbf{bold} to indicate best the result. }
\centering\label{dual_1}
\begin{tabular}{l c c }
\hline
Model & PSNR &  SSIM\\ 
\hline
Residual U-Net & 31.64 &0.9712 \\
\hline
Dual-decoder U-Net & 32.26 & 0.9724 \\ 
\hline
ADU-Net & \textbf{33.83} & \textbf{0.9784} \\ 
\hline
\end{tabular}
\end{table}

The above study shows our design flow is reasonable. We further evaluate the effectiveness of the contamination residual branch and scene residual branch in ADU-Net (see the results in Table~\ref{dual}). As compared to the Residual U-Net, each branch can improve its performance, showing the effectiveness of the proposed residual branch. Also, we can observe that the combination of the proposed residual branches can achieve further improvement, indicating that those two decoders learn complementary features of the image.

\begin{table}[!ht]
\caption{Effect of dual-branch decoder in ADU-Net. We use \textbf{bold} to indicate best the result. }
\centering\label{dual}
\begin{tabular}{l c c }
\hline
Model & PSNR &  SSIM\\ 
\hline
Residual U-Net & 31.64 &0.9712 \\
\hline
+ Contamination residual branch  & 32.30  & 0.9725 \\
\hline
+ Scene residual branch & 32.94 & 0.9744  \\ 
\hline
ADU-Net & \textbf{33.83} & \textbf{0.9784} \\ 
\hline
\end{tabular}
\end{table}

\textbf{Effect of Self-attention Module.} In this study, we evaluate the effectiveness of the self-attention mechanism in the proposed ADU-Net. The results are reported in Table~\ref{sa_1}. Table~\ref{sa_1} reveals the effectiveness of the self-attention mechanism in the proposed network. We can also observe that the W-MSA module and SW-MSA module can help the network to learn complementary information in each branch, justifying our assumption in the design. 

\begin{table}[!ht]
\caption{Effect of Self-attention Module. We use \textbf{bold} to indicate best the result. }
\centering\label{sa_1}
\begin{tabular}{l c c }
\hline
Model & PSNR &  SSIM\\ 
\hline
Dual-decoder U-Net & 32.26 & 0.9724 \\ 
\hline
+ W-MSA & 32.70 & 0.9761\\
\hline
+ SW-MSA & 32.77& 0.9760\\
\hline
+ W-MSA\&SW-MSA & \textbf{33.00}  & \textbf{0.9759} \\ 
\hline
\end{tabular}
\end{table}

\textbf{Effect of Feature Fusion Module.} In the proposed architecture of the ADU-Net, each decoder block has two information flows, respectively encoding the contamination residual and scene residual (see Fig.~\ref{dec0} and Fig.~\ref{deci}). Each information flow yields the feature fusion w.r.t. the concern of physical properties. In this study, we evaluate our design. Table~\ref{GC} ablations the effectiveness of the feature fusion blocks. Each of the CFF or GCFF can improve the accuracy by about 0.2 PSNR value. However, combining those two blocks can further bring an outstanding performance gain on top of the individual one, around 0.6 PSNR value. This can greatly verify the good practice of the feature fusion blocks in our design.

\begin{table}[!ht]
\caption{Effect of Feature Fusion Module. We use \textbf{bold} to indicate best the result. }
\centering\label{GC}
\begin{tabular}{l c c}
\hline
 Model & PSNR &  SSIM\\ \hline
Dual-decoder U-Net & 32.26 & 0.9724 \\ 
\hline   
w/o GCFF\&CFF & 33.00  & 0.9759 \\ 
\hline
+ CFF & 33.25 & 0.9770 \\ 
\hline 
+ GCFF & 33.21 & 0.9773  \\
\hline
ADU-Net & \textbf{33.83} & \textbf{0.9784} \\
\hline
\end{tabular}
\end{table}

\begin{figure*}[!ht]
\centering
\subfloat[Input]{
\includegraphics[width=0.15\linewidth]{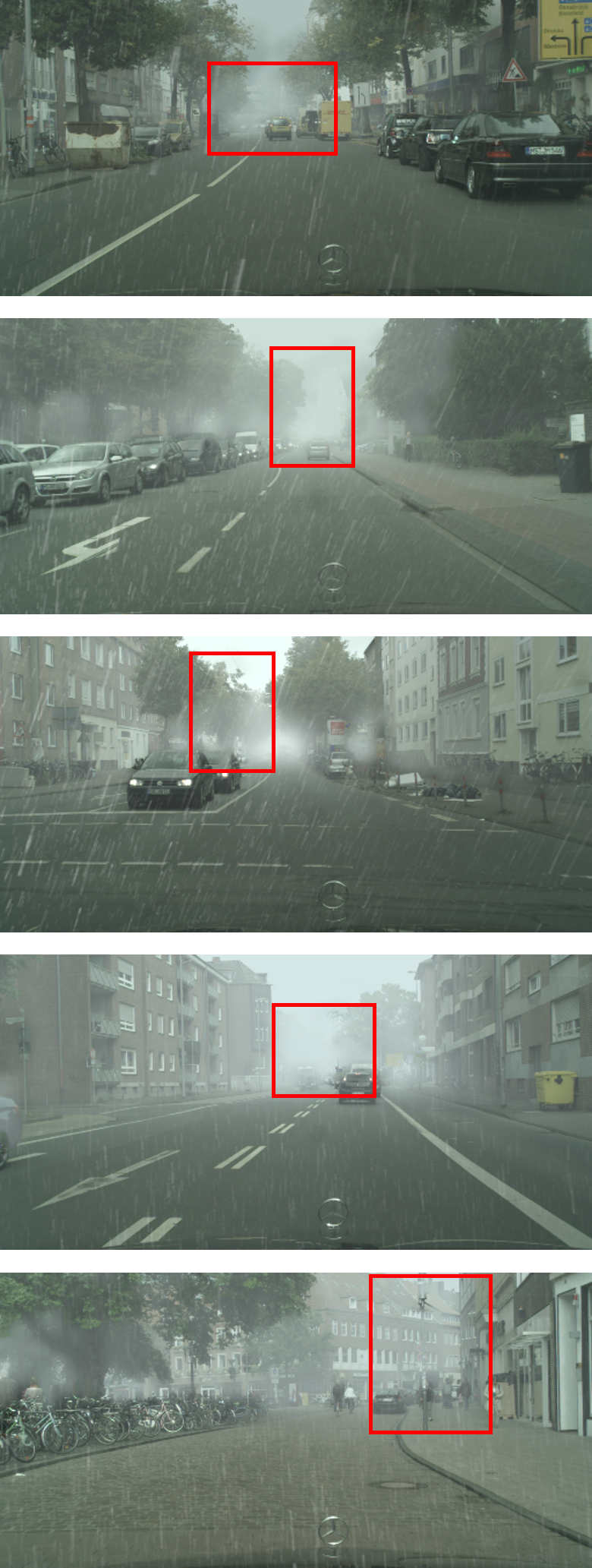}
}
\subfloat[PReNet]{
\includegraphics[width=0.15\linewidth]{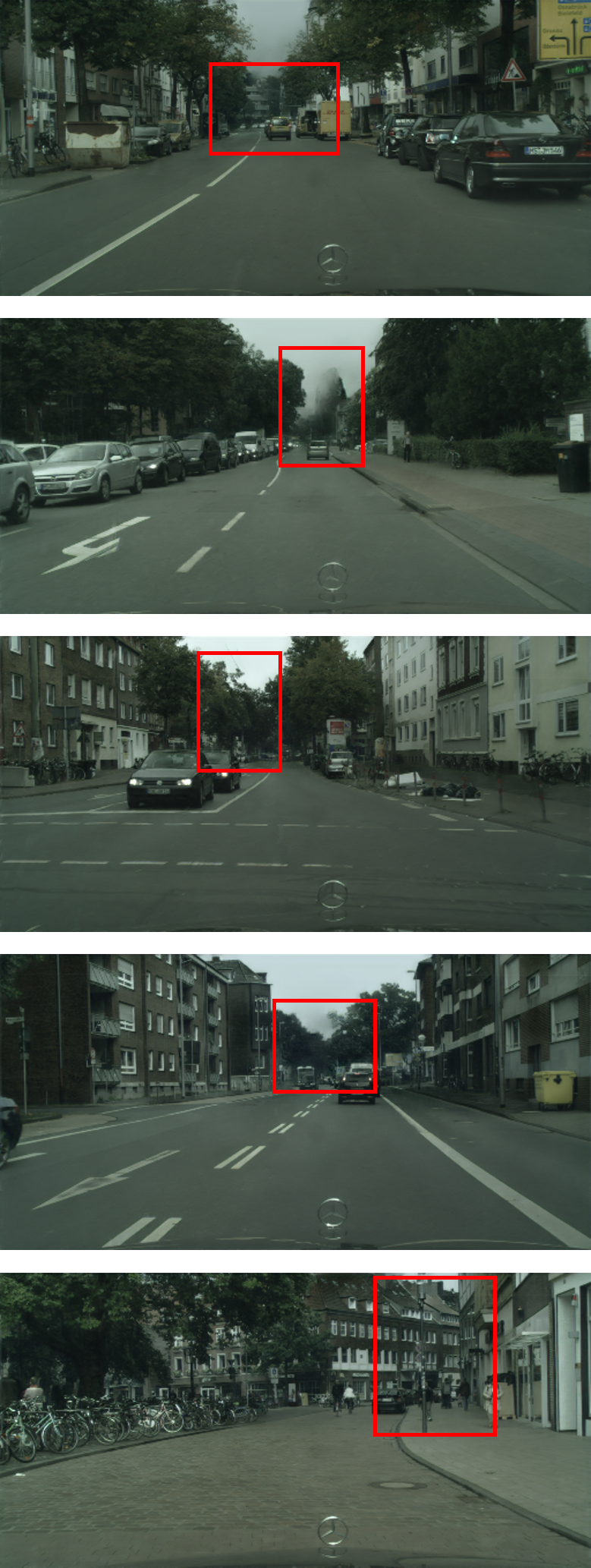}
}
\subfloat[AECR-Net]{
\includegraphics[width=0.15\linewidth]{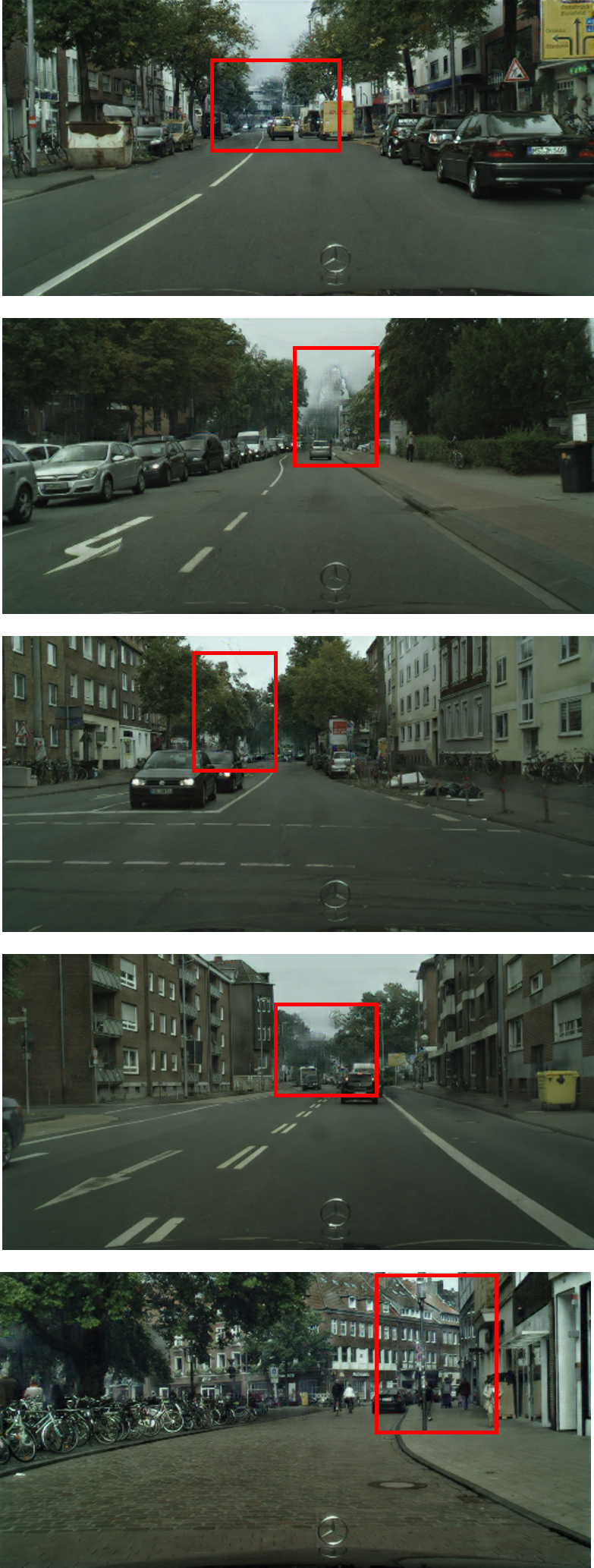}
}
\subfloat[DGNL-Net]{
\includegraphics[width=0.15\linewidth]{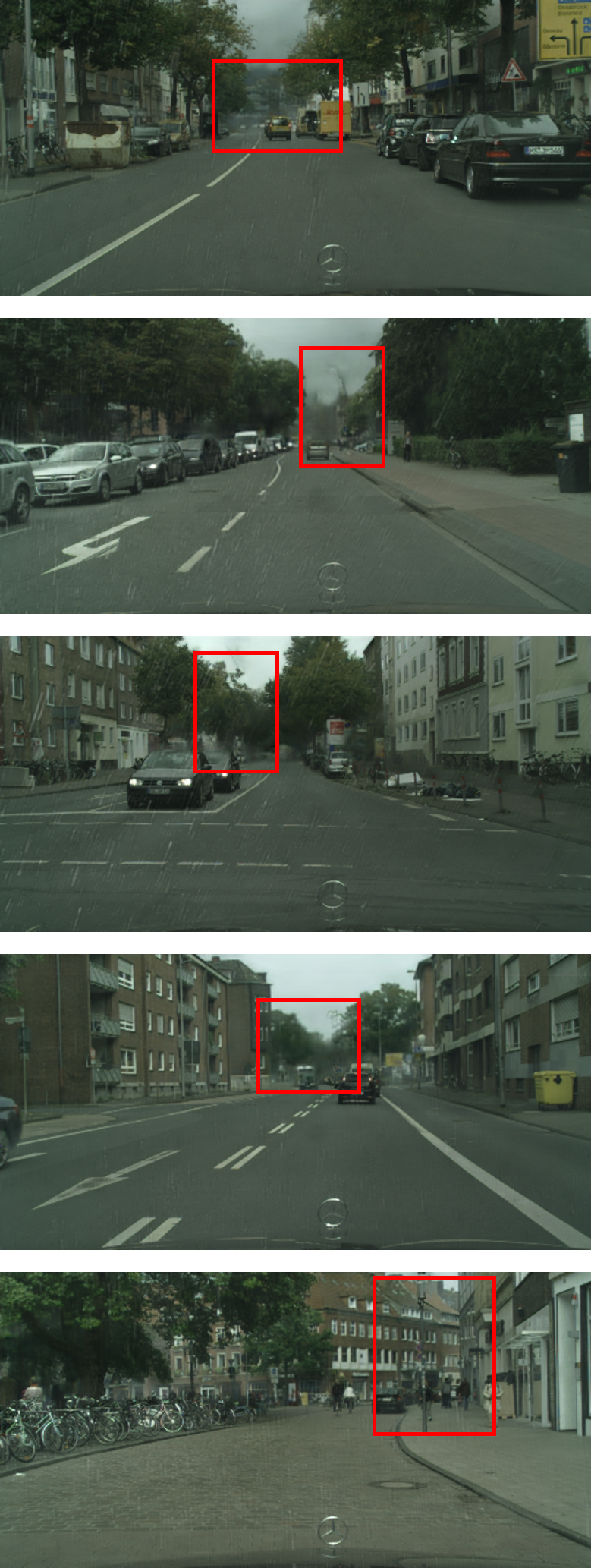}
}
\subfloat[Ours]{
\includegraphics[width=0.15\linewidth]{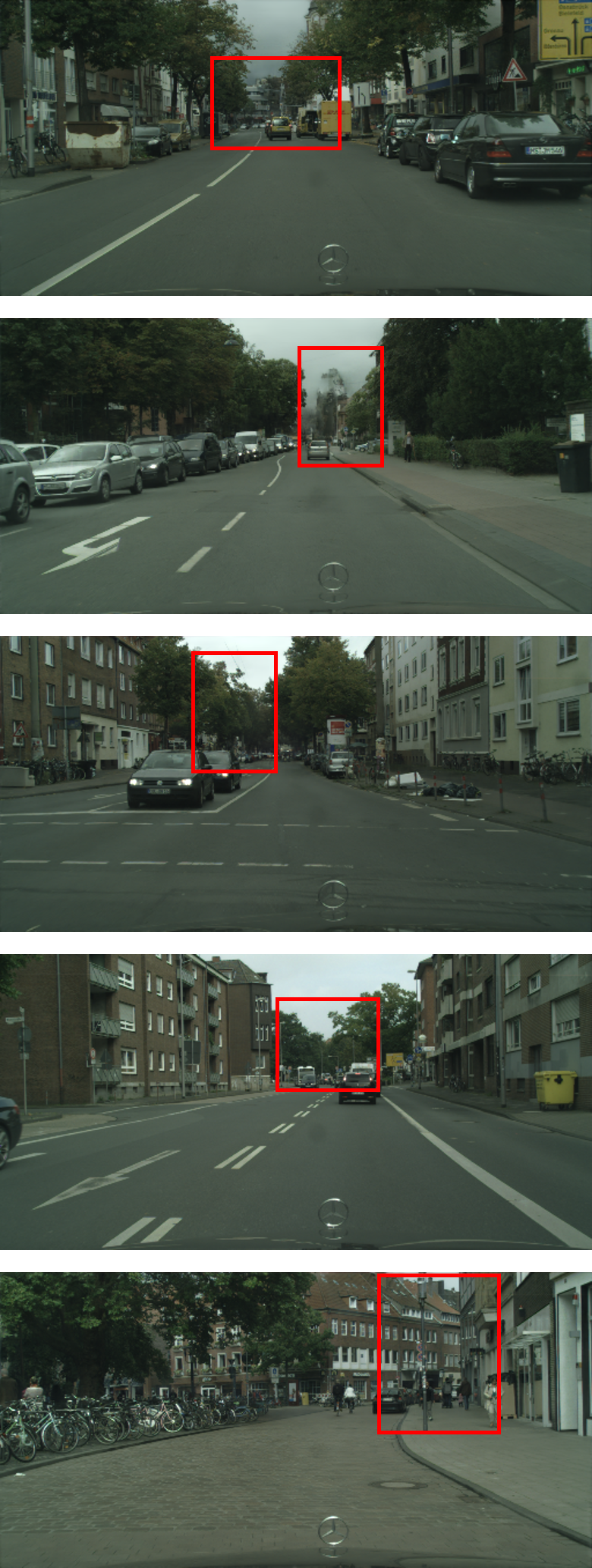}
}
\subfloat[Ground Truth]{
\includegraphics[width=0.15\linewidth]{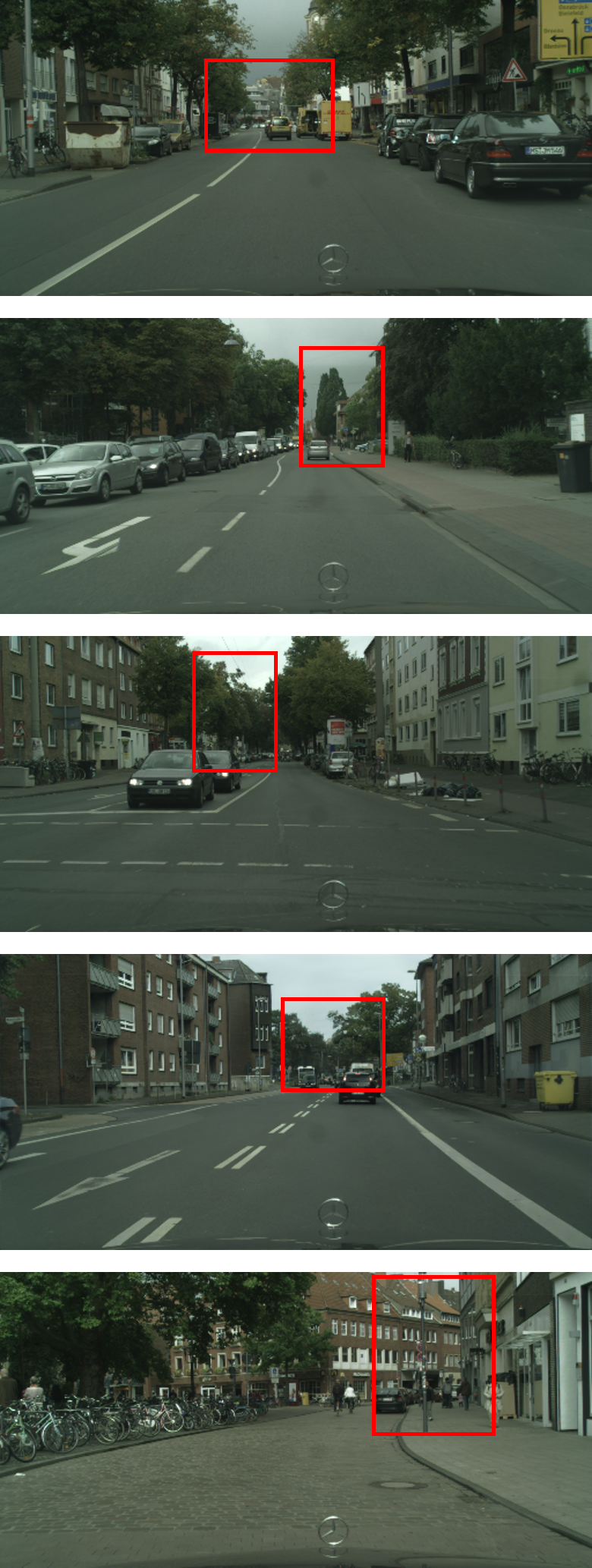}
}
\caption{Visualization of contamination removal performance on the RainCityscapes. The first column (a) is the input image. We compare our method with state-of-the-art algorithms, including PReNet~\cite{ren2019progressive}, AECR-Net~\cite{AECRNET} and DGNL-Net~\cite{dgnl2021}. (f) is the ground truth.}\label{raincity}
\end{figure*}

\begin{figure*}[!ht]
\centering
\subfloat[Input]{
\includegraphics[width=0.18\linewidth]{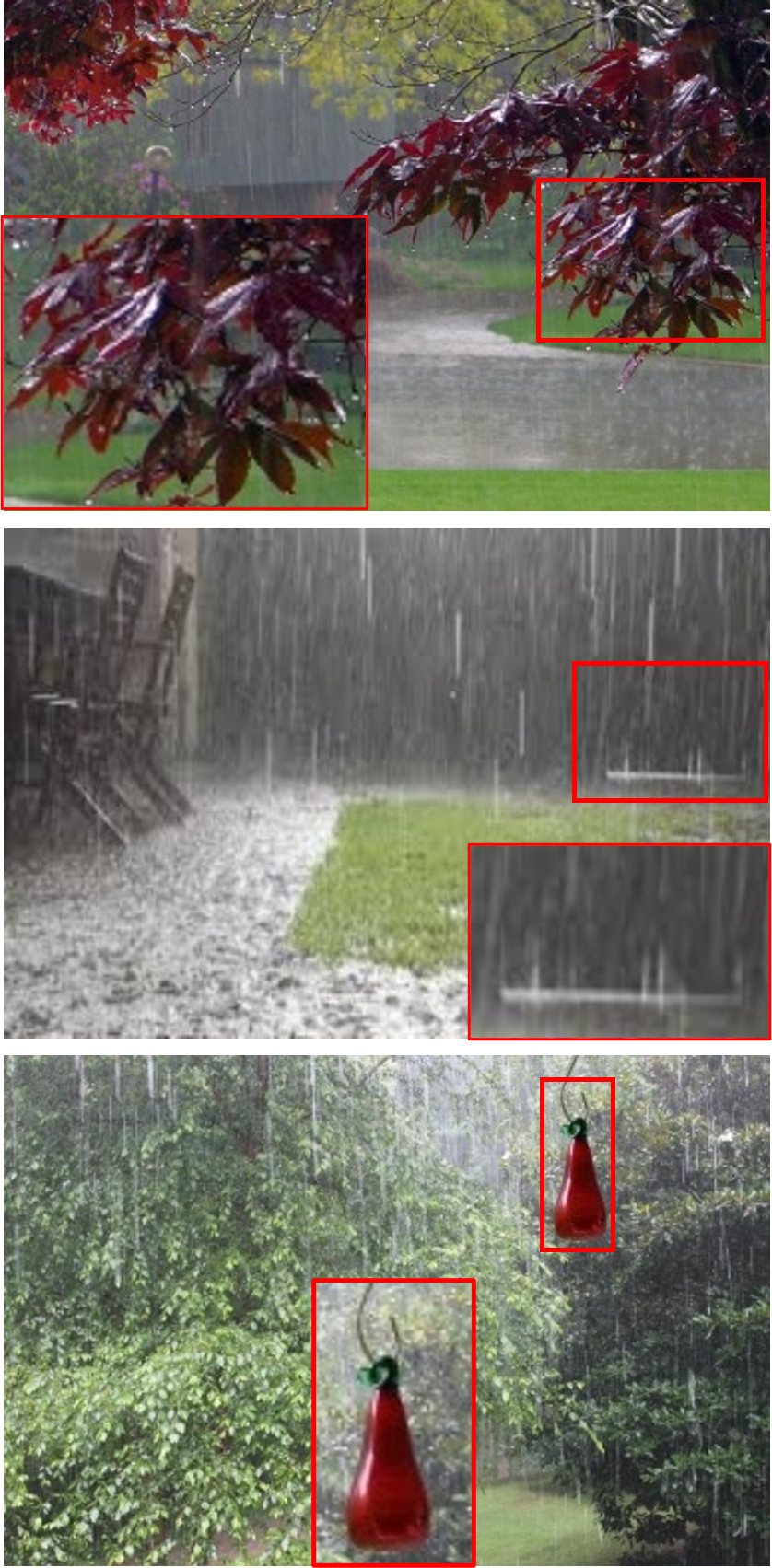}
}
\subfloat[PReNet]{
\includegraphics[width=0.18\linewidth]{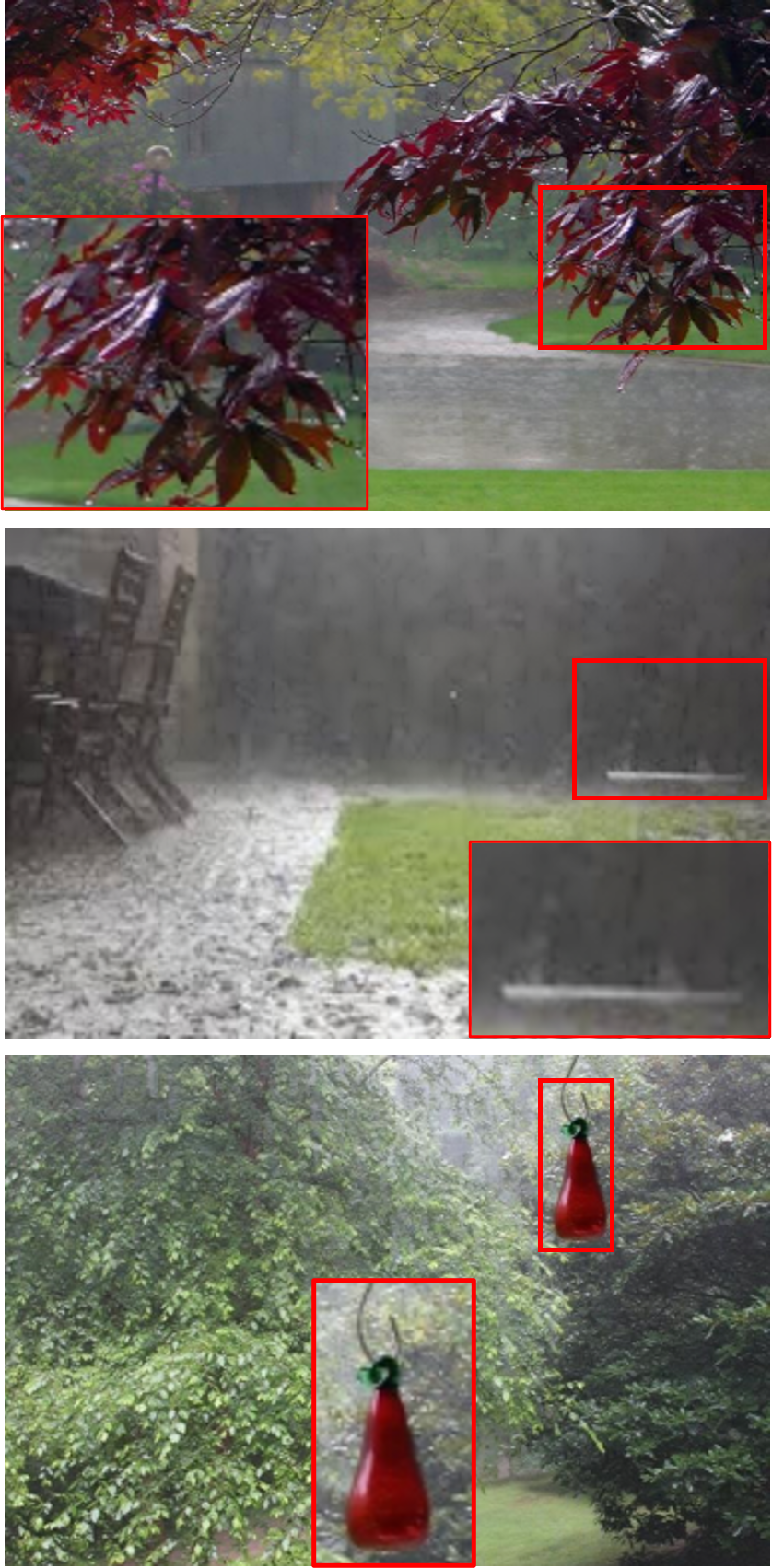}
}
\subfloat[AECR-Net]{
\includegraphics[width=0.18\linewidth]{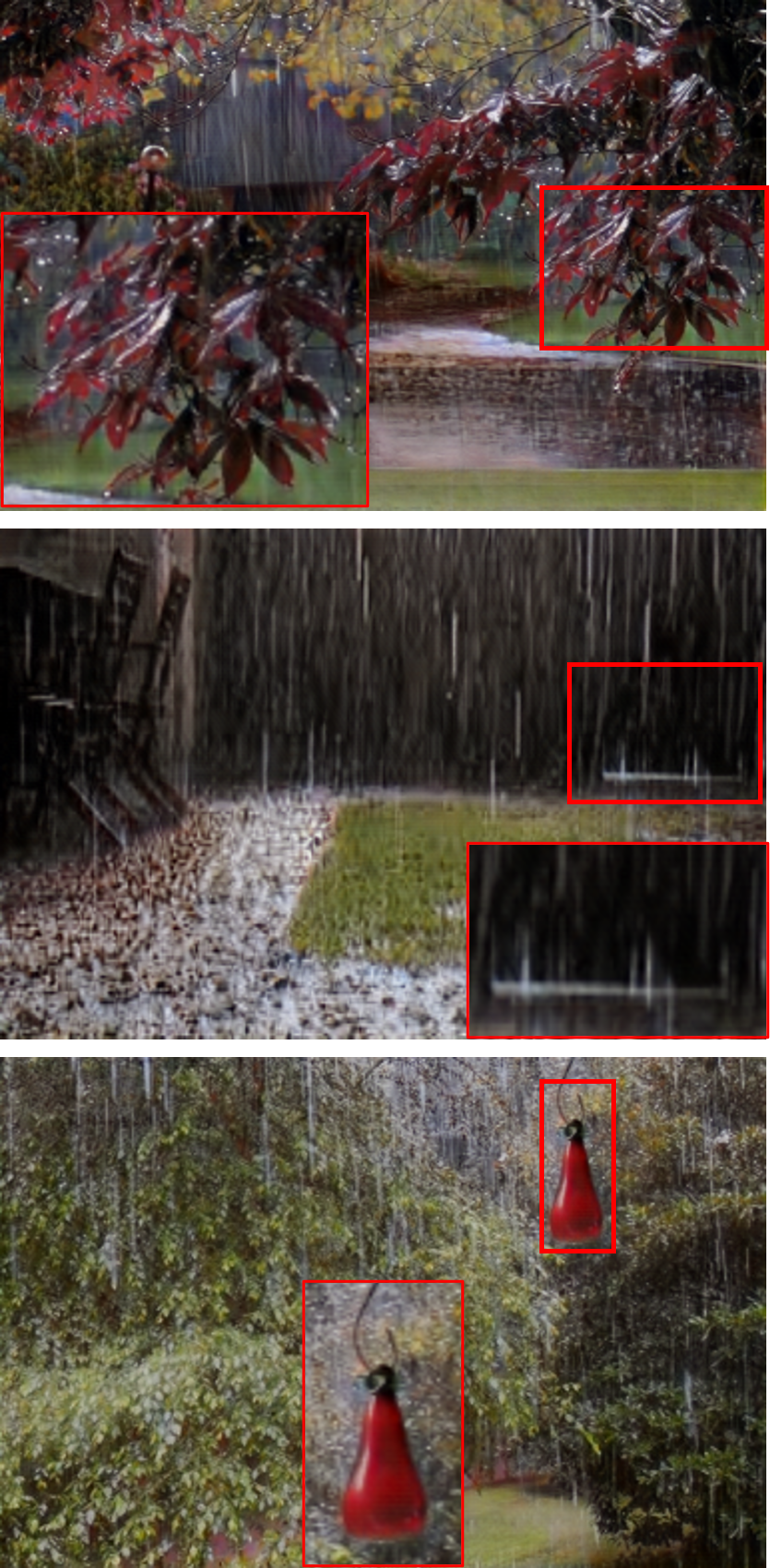}
}
\subfloat[DGNL-Net]{
\includegraphics[width=0.18\linewidth]{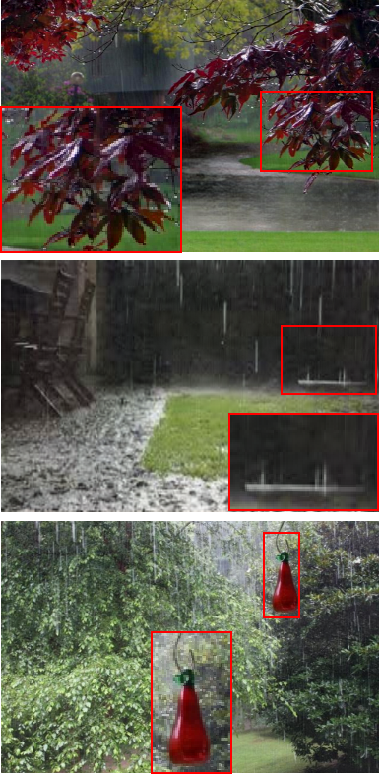}
}
\subfloat[Ours]{
\includegraphics[width=0.18\linewidth]{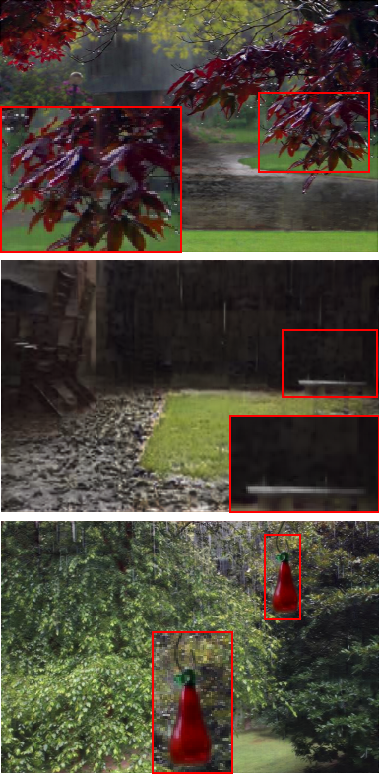}
}
\caption{Visualization of contamination removal performance on real-world images with rain and haze. The first column (a) is the input image. We compare our method with state-of-the-art algorithms, including PReNet~\cite{ren2019progressive}, AECR-Net~\cite{AECRNET} and DGNL-Net~\cite{dgnl2021}.}\label{real}
\end{figure*}

\subsection{Visualization}

\begin{figure*}[!ht]
\centering
\subfloat{
\includegraphics[width=0.18\linewidth]{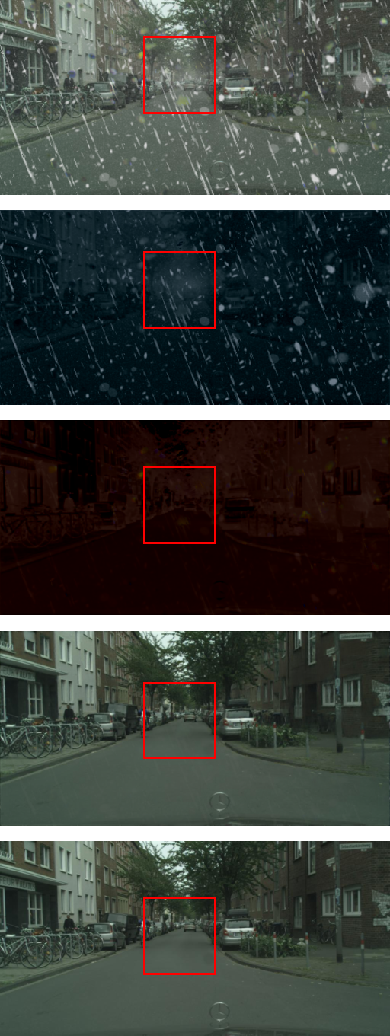}
}
\subfloat{
\includegraphics[width=0.18\linewidth]{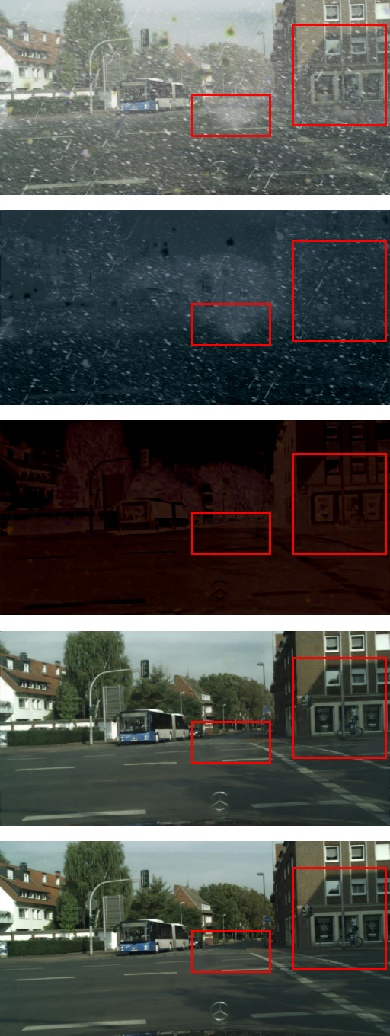}
}
\subfloat{
\includegraphics[width=0.18\linewidth]{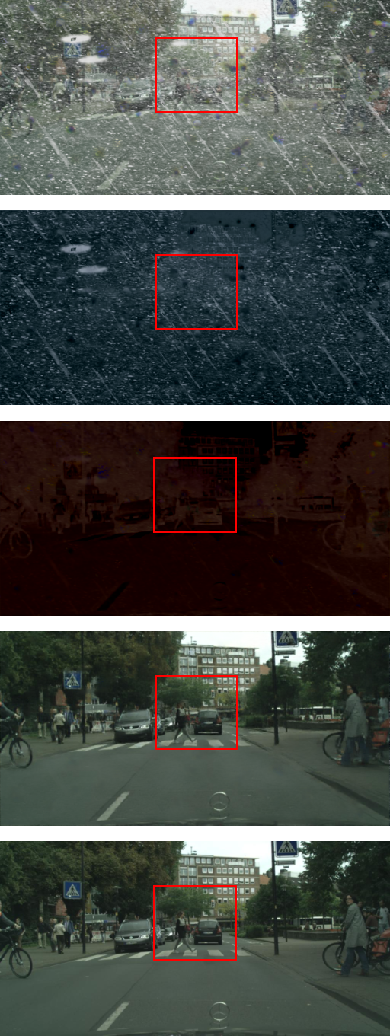}
}
\subfloat{
\includegraphics[width=0.18\linewidth]{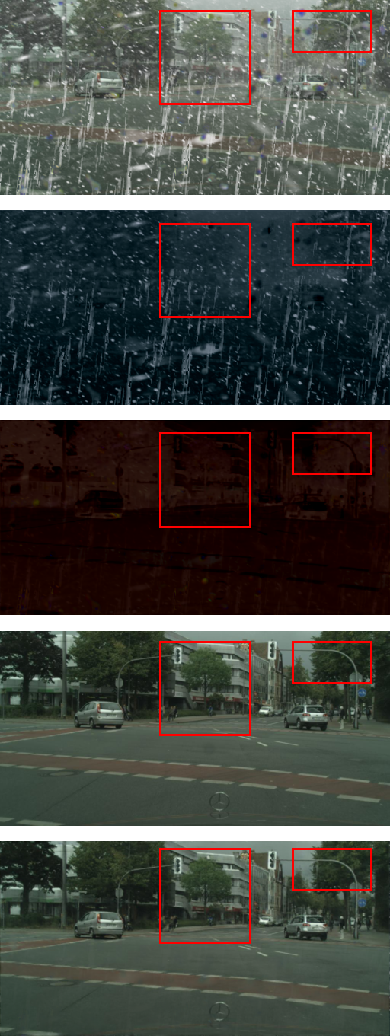}
}
\subfloat{
\includegraphics[width=0.18\linewidth]{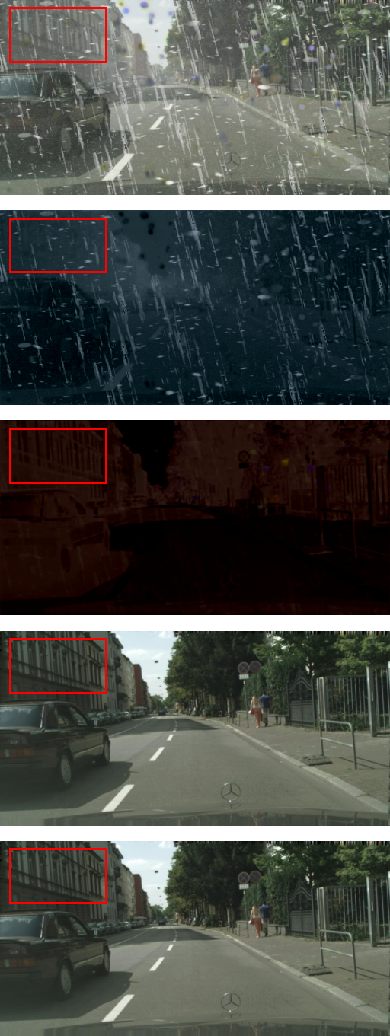}
}
\caption{Visualization of the contamination removal on the BID Rain dataset. The images in BID Rain are synthesized with rain streaks, raindrops, snow, and haze. The \textbf{first row} is the input image. The \textbf{second row} and \textbf{third row} are contamination residual and scene residual. The \textbf{fourth row} and \textbf{fifth row} are the clean image and ground truth. }\label{fout}
\end{figure*}

\begin{figure*}[!ht]
\centering
\subfloat{
\includegraphics[width=0.18\linewidth]{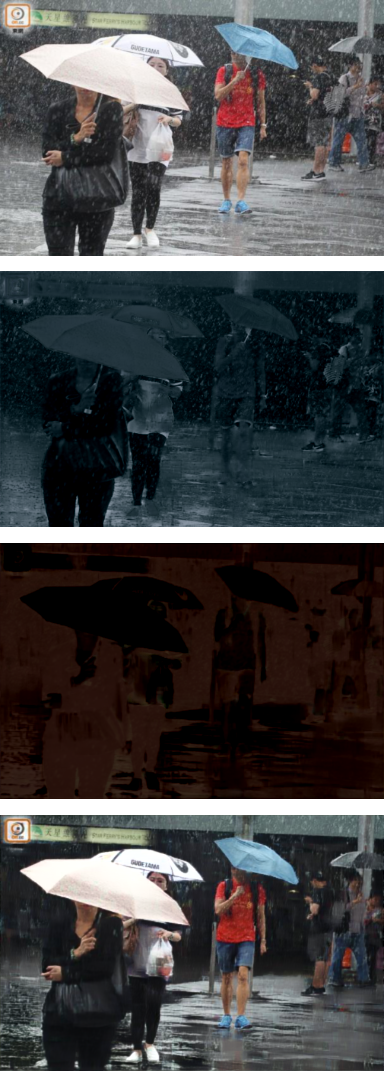}
}
\subfloat{
\includegraphics[width=0.18\linewidth]{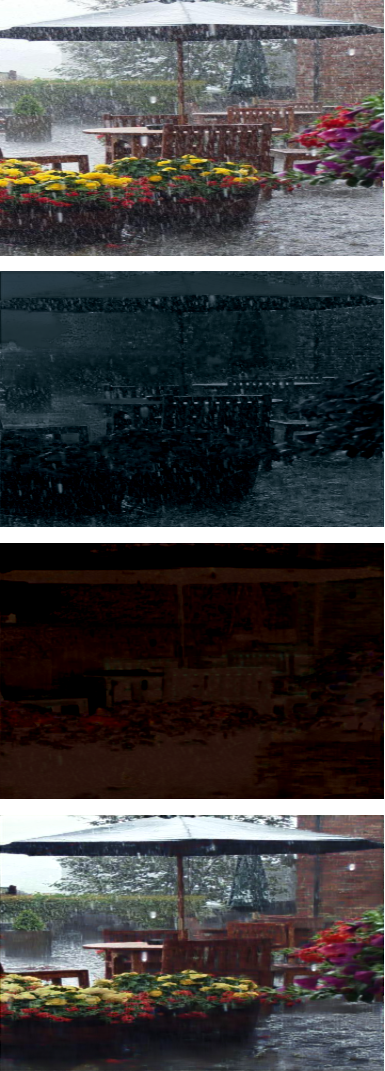}
}
\subfloat{
\includegraphics[width=0.18\linewidth]{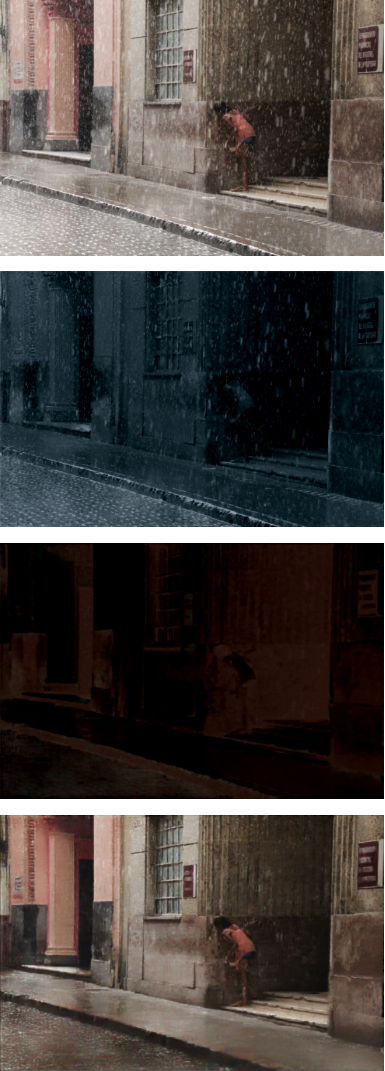}
}
\subfloat{
\includegraphics[width=0.18\linewidth]{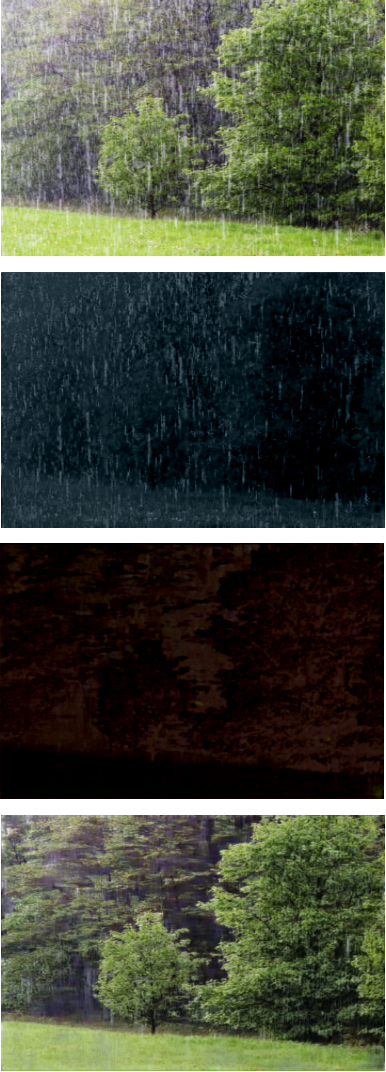}
}
\subfloat{
\includegraphics[width=0.18\linewidth]{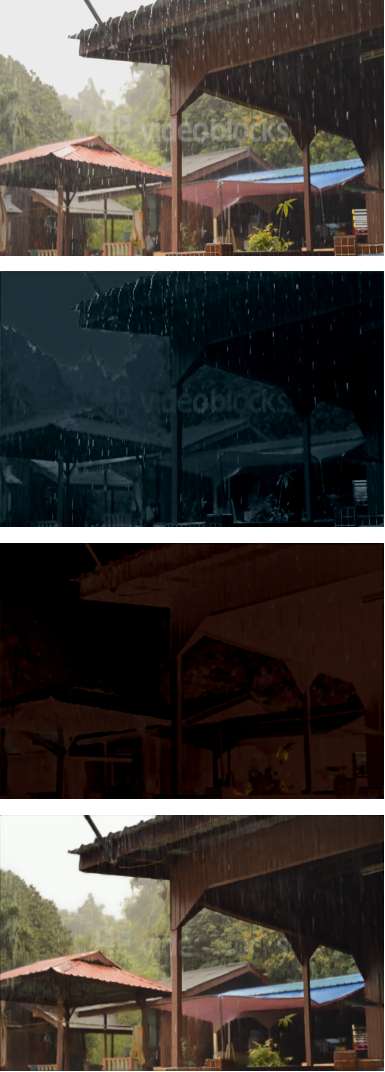}
}
\caption{Visualization of the contamination removal on real-world rain images. The \textbf{first row} is the input image. The \textbf{second row} and \textbf{third row} are contamination residual and scene residual. The \textbf{fourth row} is the clean image.}\label{real-out}
\end{figure*}

Along with the quantitative analysis in the above paragraphs, we further conduct qualitative analysis to verify the superiority of our work. In this study, we first illustrate the rain and haze removal performance between our work and existing SOTA methods in synthetic datasets (see Fig.~\ref{raincity}). Various real-world outdoor scenes are also evaluated (see Fig.~\ref{real}). The generalization of the proposed ADU-Net is further evaluated by removing other contamination, e.g. only rain in Fig.~\ref{fout}, or rain and snow in Fig.~\ref{real-out}. 

The first study is evaluated on the RainCityscapes dataset. We compare our method with the state-of-the-art methods, including PReNet~\cite{ren2019progressive}, AECR-Net~\cite{AECRNET} and DGNL-Net~\cite{dgnl2021}. As shown in Fig.~\ref{raincity}, our method can produce a much clear scene image (see the red box for details). For example, in the fourth row of Fig.~\ref{raincity}, our method removes most of the haze and produces a clear shape of the tree branches. While other methods fail to recover the tree branches. This clearly shows the superiority of our method. 

In the second study, we conduct the analysis on real-world images\footnote{147 real rain images collected from Internet.} used in~\cite{wang2019spatial}, to justify the potential of our method in real scenarios. We again compare our method to PReNet, AECR-Net, and DGNL-Net. For a fair comparison, each method adopts publicly available fine-tune weights trained on their own datasets. As can be observed from Fig.~\ref{real}, the scene images, generated by our method, are more clear and more realistic than those from other methods. For example, as compared to the rain removal network PReNet, our method can also remove the haze in real-world scenes. The hues of the recovered scene from our method are also more realistic than that from the dehazing network AECR-Net and reflective details of the scenes are maintained by our method. As compared to DGNL-Net, the closest work to ours, our ADU-Net can remove more rain streaks (the second row) or haze (the third row) and retains more scene details (the first row). This study can vividly show the effectiveness of our method in real scenarios.

To demonstrate the generalization of our dual-decoder architecture in separating different contamination, we show the residual produced by different branches. Fig.~\ref{fout} shows the results of our method on the BID Rain dataset. The first row is the input image. The second row and third row present the masks of contamination residual and scene residual. The fourth row and fifth row are the generated images and the ground truth. We can find that our method separates the contamination (e.g., snow or haze) and scene clearly, and produces high-quality scene images. A similar observation is also made in the real-world images from Internet-Data in Fig.~\ref{real-out}. This study also verifies our motivation that most of the contamination components in the image are included in the contamination residual while the scene residual contains more detail of the scene including building structures and driveway lines. This analysis again illustrates the superior generalization of the proposed method.
\section{Conclusion}
In this paper, we propose ADU-Net, the first module involving two residual branches, for the joint rain and haze removal task. Unlike previous work focusing on the contamination removal only, ADU-Net recalls the importance of restoring the scene information affected by the change of atmospheric light. By leveraging our proposed scene residual and contamination residual, ADU-Net can produce clear scene images. The superiority of ADU-Net is evaluated by extensive experiments, and the proposed ADU-Net outperforms the current state-of-the-art approaches significantly across three benchmark datasets and tasks. We believe our study will serve as a strong baseline for future work, and inspire more research work in the line of joint rain and haze removal task.


\ifCLASSOPTIONcaptionsoff
  \newpage
\fi



%
\bibliographystyle{IEEEtran}
\bibliography{TIP_bib}

\begin{thebibliography}{10}
\providecommand{\url}[1]{#1}
\csname url@samestyle\endcsname
\providecommand{\newblock}{\relax}
\providecommand{\bibinfo}[2]{#2}
\providecommand{\BIBentrySTDinterwordspacing}{\spaceskip=0pt\relax}
\providecommand{\BIBentryALTinterwordstretchfactor}{4}
\providecommand{\BIBentryALTinterwordspacing}{\spaceskip=\fontdimen2\font plus
\BIBentryALTinterwordstretchfactor\fontdimen3\font minus
  \fontdimen4\font\relax}
\providecommand{\BIBforeignlanguage}[2]{{%
\expandafter\ifx\csname l@#1\endcsname\relax
\typeout{** WARNING: IEEEtran.bst: No hyphenation pattern has been}%
\typeout{** loaded for the language `#1'. Using the pattern for}%
\typeout{** the default language instead.}%
\else
\language=\csname l@#1\endcsname
\fi
#2}}
\providecommand{\BIBdecl}{\relax}
\BIBdecl

\bibitem{Chen2019DeepIntergration}
L.~Chen, W.~Zhan, W.~Tian, Y.~He, and Q.~Zou, ``Deep integration: A multi-label
  architecture for road scene recognition,'' \emph{IEEE Transactions on Image
  Processing}, vol.~28, no.~10, pp. 4883--4898, October 2019.

\bibitem{fan2019lasot}
H.~Fan, L.~Lin, F.~Yang, P.~Chu, G.~Deng, S.~Yu, H.~Bai, Y.~Xu, C.~Liao, and
  H.~Ling, ``Lasot: A high-quality benchmark for large-scale single object
  tracking,'' in \emph{Proceedings of the IEEE/CVF Conference on Computer
  Vision and Pattern Recognition}, June 2019, pp. 5369--5378.

\bibitem{Zhang2019Co-O}
H.~Zhang, H.~Zhang, C.~Wang, and J.~Xie, ``Co-occurrent features in semantic
  segmentation,'' in \emph{Proceedings of the IEEE/CVF Conference on Computer
  Vision and Pattern Recognition}, June 2019, pp. 548--557.

\bibitem{rcdnet2020wang}
H.~Wang, Q.~Xie, Q.~Zhao, and D.~Meng, ``A model-driven deep neural network for
  single image rain removal,'' in \emph{Proceedings of the IEEE/CVF Conference
  on Computer Vision and Pattern Recognition}, June 2020, pp. 3103--3112.

\bibitem{zamir2021multi}
S.~W. Zamir, A.~Arora, S.~Khan, M.~Hayat, F.~S. Khan, M.-H. Yang, and L.~Shao,
  ``Multi-stage progressive image restoration,'' in \emph{Proceedings of the
  IEEE/CVF Conference on Computer Vision and Pattern Recognition}, June 2021,
  pp. 14\,816--14\,826.

\bibitem{AECRNET}
H.~Wu, Y.~Qu, S.~Lin, J.~Zhou, R.~Qiao, Z.~Zhang, Y.~Xie, and L.~Ma,
  ``Contrastive learning for compact single image dehazing,'' in
  \emph{Proceedings of the IEEE/CVF Conference on Computer Vision and Pattern
  Recognition}, June 2021, pp. 10\,546--10\,555.

\bibitem{dgnl2021}
X.~Hu, L.~Zhu, T.~Wang, C.-W. Fu, and P.-A. Heng, ``Single-image real-time rain
  removal based on depth-guided non-local features,'' \emph{IEEE Transactions
  on Image Processing}, vol.~30, pp. 1759--1770, January 2021.

\bibitem{FHRR2021}
D.~H. Kim, W.~J. Ahn, M.~T. Lim, T.~K. Kang, and D.~W. Kim, ``Frequency-based
  haze and rain removal network (fhrr-net) with deep convolutional
  encoder-decoder,'' \emph{Applied Sciences}, vol.~11, no.~6, March 2021.

\bibitem{blind2021han}
J.~Han, W.~Li, P.~Fang, C.~Sun, J.~Hong, M.~A. Armin, L.~Petersson, and H.~Li,
  ``Blind image decomposition,'' \emph{arXiv preprint arXiv:2108.11364}, August
  2021.

\bibitem{visionandrain}
K.~Garg and S.~K. Nayar, ``Vision and rain,'' \emph{International Journal of
  Computer Vision}, vol.~75, no.~1, p. 3–27, October 2007.

\bibitem{kang2011automatic}
L.-W. Kang, C.-W. Lin, and Y.-H. Fu, ``Automatic single-image-based rain
  streaks removal via image decomposition,'' \emph{IEEE Transactions on Image
  Processing}, vol.~21, no.~4, pp. 1742--1755, December 2012.

\bibitem{luo2015removing}
Y.~Luo, Y.~Xu, and H.~Ji, ``Removing rain from a single image via
  discriminative sparse coding,'' in \emph{IEEE International Conference on
  Computer Vision}, April 2015, pp. 3397--3405.

\bibitem{li2016rain}
Y.~Li, R.~T. Tan, X.~Guo, J.~Lu, and M.~S. Brown, ``Rain streak removal using
  layer priors,'' in \emph{Proceedings of the IEEE Conference on Computer
  Vision and Pattern Recognition}, June 2016, pp. 2736--2744.

\bibitem{he2010single}
K.~He, J.~Sun, and X.~Tang, ``Single image haze removal using dark channel
  prior,'' \emph{IEEE Transactions on Pattern Analysis and Machine
  Intelligence}, vol.~33, no.~12, pp. 2341--2353, December 2011.

\bibitem{fattal2014dehazing}
R.~Fattal, ``Dehazing using color-lines,'' \emph{ACM Transactions on Graphics},
  vol.~34, no.~1, pp. 1--14, November 2014.

\bibitem{zhu2015fast}
Q.~Zhu, J.~Mai, and L.~Shao, ``A fast single image haze removal algorithm using
  color attenuation prior,'' \emph{IEEE Transactions on Image Processing},
  vol.~24, no.~11, pp. 3522--3533, November 2015.

\bibitem{zhu2017joint}
L.~Zhu, C.-W. Fu, D.~Lischinski, and P.-A. Heng, ``Joint bi-layer optimization
  for single-image rain streak removal,'' in \emph{Proceedings of the IEEE
  International Conference on Computer Vision}, October 2017, pp. 2545--2553.

\bibitem{chinese2021}
C.~LIANG, Y.~FENG, H.~XIE, M.~WEI, and X.~YAN, ``Prior-based single image rain
  and haze removal,'' \emph{Journal of ZheJiang University (Science Edition)},
  vol.~48, no.~3, pp. 270--281, May 2021.

\bibitem{li2017aod}
B.~Li, X.~Peng, Z.~Wang, J.~Xu, and D.~Feng, ``Aod-net: All-in-one dehazing
  network,'' in \emph{Proceedings of the IEEE International Conference on
  Computer Vision}, October 2017, pp. 4780--4788.

\bibitem{ren2020single}
W.~Ren, J.~Pan, H.~Zhang, X.~Cao, and M.-H. Yang, ``Single image dehazing via
  multi-scale convolutional neural networks with holistic edges,''
  \emph{International Journal of Computer Vision}, vol. 128, no.~1, pp.
  240--259, January 2020.

\bibitem{chen2019pms}
W.-T. Chen, J.-J. Ding, and S.-Y. Kuo, ``Pms-net: Robust haze removal based on
  patch map for single images,'' in \emph{Proceedings of the IEEE/CVF
  Conference on Computer Vision and Pattern Recognition}, June 2019, pp.
  11\,673--11\,681.

\bibitem{zhang2019image}
H.~Zhang, V.~Sindagi, and V.~M. Patel, ``Image de-raining using a conditional
  generative adversarial network,'' \emph{IEEE Transactions on Circuits and
  Systems for Video Technology}, vol.~30, no.~11, pp. 3943--3956, November
  2020.

\bibitem{ren2019progressive}
D.~Ren, W.~Zuo, Q.~Hu, P.~Zhu, and D.~Meng, ``Progressive image deraining
  networks: A better and simpler baseline,'' in \emph{Proceedings of the
  IEEE/CVF Conference on Computer Vision and Pattern Recognition}, June 2019,
  pp. 3932--3941.

\bibitem{wang2019spatial}
T.~Wang, X.~Yang, K.~Xu, S.~Chen, Q.~Zhang, and R.~W. Lau, ``Spatial attentive
  single-image deraining with a high quality real rain dataset,'' in
  \emph{Proceedings of the IEEE/CVF Conference on Computer Vision and Pattern
  Recognition}, June 2019, pp. 12\,262--12\,271.

\bibitem{fu2017removing}
X.~Fu, J.~Huang, D.~Zeng, Y.~Huang, X.~Ding, and J.~Paisley, ``Removing rain
  from single images via a deep detail network,'' in \emph{Proceedings of the
  IEEE Conference on Computer Vision and Pattern Recognition}, July 2017, pp.
  1715--1723.

\bibitem{yang2017deep}
W.~Yang, R.~T. Tan, J.~Feng, J.~Liu, Z.~Guo, and S.~Yan, ``Deep joint rain
  detection and removal from a single image,'' in \emph{Proceedings of the IEEE
  Conference on Computer Vision and Pattern Recognition}, July 2017, pp.
  1685--1694.

\bibitem{li2018non}
G.~Li, X.~He, W.~Zhang, H.~Chang, L.~Dong, and L.~Lin, ``Non-locally enhanced
  encoder-decoder network for single image de-raining,'' in \emph{Proceedings
  of the 26th ACM international conference on Multimedia}, October 2018, pp.
  1056--1064.

\bibitem{li2018recurrent}
X.~Li, J.~Wu, Z.~Lin, H.~Liu, and H.~Zha, ``Recurrent squeeze-and-excitation
  context aggregation net for single image deraining,'' in \emph{Proceedings of
  the European Conference on Computer Vision}, October 2018, pp. 262--277.

\bibitem{zhang2018density}
H.~Zhang and V.~M. Patel, ``Density-aware single image de-raining using a
  multi-stream dense network,'' in \emph{Proceedings of the IEEE Conference on
  Computer Vision and Pattern Recognition}, June 2018, pp. 695--704.

\bibitem{isola2017image}
P.~Isola, J.-Y. Zhu, T.~Zhou, and A.~A. Efros, ``Image-to-image translation
  with conditional adversarial networks,'' in \emph{Proceedings of the IEEE
  Conference on Computer Vision and Pattern Recognition}, July 2017, pp.
  5967--5976.

\bibitem{zhu2020learning}
L.~Zhu, Z.~Deng, X.~Hu, H.~Xie, X.~Xu, J.~Qin, and P.-A. Heng, ``Learning gated
  non-local residual for single-image rain streak removal,'' \emph{IEEE
  Transactions on Circuits and Systems for Video Technology}, vol.~31, no.~6,
  pp. 2147--2159, June 2021.

\bibitem{cai2016dehazenet}
B.~Cai, X.~Xu, K.~Jia, C.~Qing, and D.~Tao, ``Dehazenet: An end-to-end system
  for single image haze removal,'' \emph{IEEE Transactions on Image
  Processing}, vol.~25, no.~11, pp. 5187--5198, November 2016.

\bibitem{ren2016single}
W.~Ren, S.~Liu, H.~Zhang, J.~Pan, X.~Cao, and M.-H. Yang, ``Single image
  dehazing via multi-scale convolutional neural networks,'' in \emph{European
  Conference on Computer Vision}.\hskip 1em plus 0.5em minus 0.4em\relax
  Springer, September 2016, pp. 154--169.

\bibitem{zhang2018densely}
H.~Zhang and V.~M. Patel, ``Densely connected pyramid dehazing network,'' in
  \emph{Proceedings of the IEEE Conference on Computer Vision and Pattern
  Recognition}, June 2018, pp. 3194--3203.

\bibitem{ren2018gated}
W.~Ren, L.~Ma, J.~Zhang, J.~Pan, X.~Cao, W.~Liu, and M.-H. Yang, ``Gated fusion
  network for single image dehazing,'' in \emph{Proceedings of the IEEE
  Conference on Computer Vision and Pattern Recognition}, June 2018, pp.
  3253--3261.

\bibitem{qu2019enhanced}
Y.~Qu, Y.~Chen, J.~Huang, and Y.~Xie, ``Enhanced pix2pix dehazing network,'' in
  \emph{Proceedings of the IEEE/CVF Conference on Computer Vision and Pattern
  Recognition}, June 2019, pp. 8152--8160.

\bibitem{ssim}
Z.~Wang, A.~Bovik, H.~Sheikh, and E.~Simoncelli, ``Image quality assessment:
  from error visibility to structural similarity,'' \emph{IEEE Transactions on
  Image Processing}, vol.~13, no.~4, pp. 600--612, April 2004.

\bibitem{swin}
Z.~Liu, Y.~Lin, Y.~Cao, H.~Hu, Y.~Wei, Z.~Zhang, S.~Lin, and B.~Guo, ``Swin
  transformer: Hierarchical vision transformer using shifted windows,'' in
  \emph{IEEE/CVF International Conference on Computer Vision}, October 2021,
  pp. 9992--10\,002.

\bibitem{chen2021robust}
C.~Chen and H.~Li, ``Robust representation learning with feedback for single
  image deraining,'' in \emph{IEEE/CVF Conference on Computer Vision and
  Pattern Recognition}, June 2021, pp. 7738--7747.

\bibitem{BatchNA}
S.~Ioffe and C.~Szegedy, ``Batch normalization: Accelerating deep network
  training by reducing internal covariate shift,'' in \emph{Proceedings of the
  32nd International Conference on International Conference on Machine
  Learning}, ser. ICML'15, vol.~37.\hskip 1em plus 0.5em minus 0.4em\relax
  JMLR.org, July 2015, p. 448–456.

\bibitem{relu}
A.~F. Agarap, ``Deep learning using rectified linear units (relu),''
  \emph{arXiv preprint arXiv:1803.08375}, March 2018.

\bibitem{pytorch}
A.~Paszke, S.~Gross, F.~Massa, A.~Lerer, J.~Bradbury, G.~Chanan, T.~Killeen,
  Z.~Lin, N.~Gimelshein, L.~Antiga, A.~Desmaison, A.~Kopf, E.~Yang, Z.~DeVito,
  M.~Raison, A.~Tejani, S.~Chilamkurthy, B.~Steiner, L.~Fang, J.~Bai, and
  S.~Chintala, ``Pytorch: An imperative style, high-performance deep learning
  library,'' in \emph{Advances in Neural Information Processing Systems
  32}.\hskip 1em plus 0.5em minus 0.4em\relax Curran Associates, Inc., December
  2019, pp. 8024--8035.

\bibitem{cityscape}
M.~Cordts, M.~Omran, S.~Ramos, T.~Rehfeld, M.~Enzweiler, R.~Benenson,
  U.~Franke, S.~Roth, and B.~Schiele, ``The cityscapes dataset for semantic
  urban scene understanding,'' in \emph{Proceedings of the IEEE Conference on
  Computer Vision and Pattern Recognition}, June 2016, pp. 3213--3223.

\bibitem{snow100}
Y.-F. Liu, D.-W. Jaw, S.-C. Huang, and J.-N. Hwang, ``Desnownet: Context-aware
  deep network for snow removal,'' \emph{IEEE Transactions on Image
  Processing}, vol.~27, no.~6, pp. 3064--3073, June 2018.

\bibitem{foggycityscape}
C.~Sakaridis, D.~Dai, and L.~V. Gool, ``Semantic foggy scene understanding with
  synthetic data,'' \emph{International Journal of Computer Vision}, vol. 126,
  pp. 973--992, September 2018.

\bibitem{metaball}
J.~F. Blinn, ``A generalization of algebraic surface drawing,'' \emph{ACM
  Transactions on Graphic}, vol.~1, no.~3, p. 235–256, July 1982.

\bibitem{liu2019dual}
X.~Liu, M.~Suganuma, Z.~Sun, and T.~Okatani, ``Dual residual networks
  leveraging the potential of paired operations for image restoration,'' in
  \emph{Proceedings of the IEEE/CVF Conference on Computer Vision and Pattern
  Recognition}, June 2019, pp. 7000--7009.

\bibitem{dafhu2019depth}
X.~Hu, C.-W. Fu, L.~Zhu, and P.-A. Heng, ``Depth-attentional features for
  single-image rain removal,'' in \emph{Proceedings of the IEEE/CVF Conference
  on Computer Vision and Pattern Recognition}, June 2019, pp. 8014--8023.

\bibitem{ssim+mse}
Z.~Fan, H.~Wu, X.~Fu, Y.~Huang, and X.~Ding, ``Residual-guide network for
  single image deraining,'' in \emph{Proceedings of the 26th ACM International
  Conference on Multimedia}.\hskip 1em plus 0.5em minus 0.4em\relax Association
  for Computing Machinery, October 2018, p. 1751–1759.

\end{thebibliography}


\end{document}